\newcommand*{\ARXIV}{}  % /!\ CHANGE compiler to TexLive 2017!!!

\ifdefined\IEEE
    \documentclass[journal]{IEEEtran}
    \usepackage{amsmath,amsfonts}
    \usepackage{algorithmic}
    \usepackage{algorithm}
    \usepackage{array}
    \usepackage[caption=false,font=normalsize,labelfont=sf,textfont=sf]{subfig}
    \usepackage{textcomp}
    \usepackage{stfloats}
    \usepackage{url}
    \usepackage{verbatim}
    \usepackage{graphicx}
    \usepackage{cite}
    \usepackage[numbers]{natbib}
\fi

\ifdefined\ARXIV
    \documentclass{article}
    \usepackage[utf8]{inputenc}
    \usepackage{arxiv}
    \usepackage{moreverb,url}
    \usepackage{amssymb}
    \usepackage{caption}
    \usepackage{subcaption}
    \usepackage{amsmath}
    
    \newcommand\BibTeX{{\rmfamily B\kern-.05em \textsc{i\kern-.025em b}\kern-.08em
    T\kern-.1667em\lower.7ex\hbox{E}\kern-.125emX}}
\fi

\newcommand\figref{Fig.~\ref}
\newcommand\secref{Sec.~\ref}
\newcommand\tabref{Tab.~\ref}
\newcommand\equref{Eq.~\ref}

\usepackage{multirow}
\usepackage{graphicx}
\usepackage{booktabs}

\ifdefined\ARXIV
    \usepackage[style=nature, backend=bibtex, natbib=true, hyperref = true, url=false, isbn=false, doi=false]{biblatex}
    \usepackage{graphicx}
    \usepackage{hyperref}
    \usepackage{comment}
    \hypersetup{colorlinks,allcolors=black}
\fi

\usepackage{todonotes}
\usepackage{xcolor}
\newcommand{\titleName}{Learning Informative Health Indicators Through Unsupervised Contrastive Learning}
\newcommand{\abstr}{% THE PROBLEM
{\color{black}Monitoring the health of complex industrial assets is crucial for safe and efficient operations. Health indicators that provide  quantitative real-time insights into the health status of industrial assets over time serve as valuable tools for e.g.\ fault detection or prognostics. This study proposes a novel, versatile and unsupervised approach to learn health indicators using contrastive learning, where the \textit{operational time} serves as a proxy for degradation. To highlight its versatility, the approach is evaluated on two tasks and case studies with different characteristics: wear assessment of milling machines and fault detection of railway wheels. Our results show that the proposed methodology effectively learns a health indicator that follows the wear of  milling machines (0.97 correlation on average) and is suitable for fault detection in railway wheels ($\mathbf{88.7\%}$ balanced accuracy). The conducted experiments demonstrate the versatility of the approach for various systems and health conditions.}

}

\begin{document}
\ifdefined\IEEE
\title{\titleName}

\author{%
        Katharina Rombach \thanks{Katharina Rombach is at the Chair of Intelligent Maintenance Systems, ETH Zurich} 
        Gabriel Michau \thanks{
        Dr. Gabriel Michau is at the Maintenance Systems \& Technologies group,
        Stadler Service AG}
        Wilfried Bürzle \thanks{
        Dr. Wilfried Bürzle is at Vehicle and System Engineering group,
        Swiss Federal Railways}
        Stefan Koller 
        \thanks{Dr. Stefan Koller is at the Wayside Train Monitoring Systems Department group,
        Swiss Federal Railways}
        Olga Fink \IEEEmembership{member, IEEE}
        \thanks{Olga Fink is the Professor of the Intelligent Maintenance and Operations Systems at
        EPFL}
        % <-this % stops a space
}

\markboth{IEEE Transactions on Reliability}%
{}
        %\date{\today}

\maketitle
\begin{abstract}
\abstr
\end{abstract}

\begin{IEEEkeywords}
        Unsupervised Health Indicator, Unsupervised Contrastive Learning, Fault Detection, Wear Assessment
        \end{IEEEkeywords}
\fi

\ifdefined\ARXIV
    \title{\titleName}
    \author{%
        Katharina Rombach\\
        \textit{\small{Chair of Intelligent}} \\\textit{\small{Maintenance Systems}}\\
        ETH Z\"urich,\\
        Z\"urich, Switzerland\\
        \And 
        Gabriel Michau\\
        \textit{\small{Maintenance Systems}} \\\textit{\small{\& Technologies}}\\
        Stadler Service AG,\\
        Bussnang, Switzerland\\
        \And 
        Dr. Wilfried Bürzle \\
        \textit{\small{Vehicle and }} \\\textit{\small{  System Engineering,}}\\
        Swiss Federal Railways,\\
        Switzerland.\\
        \And 
        Dr. Stefan Koller \\
        \textit{\small{Wayside Train Monitoring  }} \\\textit{\small{ Systems Department,}}\\
        Swiss Federal Railways,\\
        Switzerland.\\
        \And 
        Olga Fink\\
        \textit{\small{Intelligent Maintenance}} \\\textit{\small{and Operations Systems}}\\
        EPFL,\\
        Lausanne, Switzerland}
        %\subtitle{Preprint}
        \date{\today}
        \maketitle
        \begin{abstract}
        \abstr
        \end{abstract}
        \keywords{Unsupervised Health Indicator, Unsupervised Contrastive Learning, Anomaly Detection, Wear Assessment}
\fi

\section{Introduction} \label{sec:Introduction}
\ifdefined\IEEE
\IEEEPARstart{C}{ondition} monitoring (CM) plays a crucial role for the safe, reliable and efficient operation of complex industrial assets.
\fi
\ifdefined\ARXIV
Condition monitoring (CM) plays a crucial role for the safe, reliable and efficient operation of complex industrial assets.
\fi
The health condition of an asset is usually described by a health indicator or a condition indicator, which reflect the health status in a predictable way~\cite{lei2018machinery, garmaev2023deep}.
Condition indicators can be any of the features within a system that enable to differentiate degradation states or normal operation from faulty conditions in a predictable way \cite{fink2020potential}. 
Contrary to condition indicators, health indicators integrate  several inputs or multiple condition indicators into a single value, providing the end user with an aggregated health status of the component \cite{fink2020potential}.
Formulating precise requirements for
health indicators is challenging, however, some common characteristics
can be highlighted: detectability, trendability, prognosability, and monotonicity \cite{garmaev2023deep, lei2018machinery}. {\color{black}These characteristics can only be met under ideal circumstances. If only parts of the asset are monitored, for example, some measurements might not observe a local defect. Due to the partial observability, the health indicator can seemingly recover if a defect is not captured by one of the sensors.  In this scenario, monotonicity cannot be requested.   Similarly, if only parts of the system such as railway wheels are monitored, defects might propagate through a system and might not be visible anymore in the available monitoring data over time. This has been observed e.g.\ for railway wheels \cite{rail}. }

Optimally, a health indicator 
can form a foundation that connects various  monitoring tasks, including fault detection, degradation assessment and prognosis applications \cite{zhou2022construction}.
Consequently, health indicators are a powerful tool for condition monitoring as they can be applied to prevent catastrophic failures on the one hand, and to avoid unnecessary time- and cost- consuming maintenance actions on the other hand.  

%SOTA
In literature, different approaches have been proposed to construct condition and health indicators of a system.
Statistical parameters such as the root mean square error (RMS) or the kurtosis have been widely applied for monitoring the condition of industrial assets \cite{zhang2023health}.
However, these condition indicators might suffer from incomplete information and thus, not represent the health condition sufficiently \cite{zhou2022construction}.
Features extracted by signal processing methods have also been proposed as condition indicators to monitor industrial assets over time \cite{firla2016automatic, saidi2017wind}.
These methods aim to enhance the health-state-related component
of the monitored signal by signal preprocessing \cite{zhou2022construction}. 
For example, \citet{firla2016automatic} proposed a complex signal filtering method (multi-rate filtering) and used the knowledge about domain specific fault characteristic frequency to 
calculate the preprocessing-based health indicator for bearing CM.
However, signal preprocessing methods require {\color{black}substantial} application specific domain knowledge{\color{black}. Not only the correct preprocessing method needs to be chosen} to be able to enhance the relevant components in the data {\color{black} but also the relevant features need to be observed. T}hus, {\color{black} these methods are often specific to applications and not as versatile}. 
To construct a less application specific health indicator that monitors the condition of the entire systems and is applicable {\color{black}with less} domain-specific knowledge, several research studies have proposed machine learning methods for learning health indicators \cite{stetco2019machine}. 
One-class classification-based (OCC-based) methods \cite{sadooghi2018improving, michau2020feature} as well as  signal reconstruction methods have been proposed in literature \cite{malhotra2016multi, ye2021health, hsu2023comparison}. 
Trained on only healthy data, both types of methods are particularly suited for condition monitoring tasks as they do not require any faulty samples. 
While OCC-based methods are often used for fault detection as they provide binary outputs (healthy or unhealthy) \cite{michau2020feature}, the measured
distance to the healthy data can be interpreted as a
health indicator. Signal reconstruction based methods train autoencoders 
to reconstruct the healthy data of the asset and the health indicator is then defined as the distance to healthy data \cite{fink2020potential}. Different approaches can then be applied to aggregate the residuals into a single health indicator \cite{wang2017prognostics}. 
Although theoretically, machine learning approaches are less application specific as e.g.\ {\color{black} pure} preprocessing techniques, some approaches still require some domain knowledge or are only evaluated on one type of application
(e.g.\ gearbox, shafts, bearings, batteries, etc.) \cite{atamuradov2020machine, garmaev2023deep}.
Furthermore, health indicators are often developed for only one of the CM tasks (detection, wear assessment or prognostics). For example, health indicators that are achieving
detectability and separability are optimal for
fault detection whereas trendability indicates the optimal
health indicator for describing the degradation trend \cite{zhou2022construction}. 
 Optimal health indicators, however, should be {\color{black}versatilely} applicable i.a.\ suited to all of the above mentioned CM tasks and to different applications.
In addition, despite the recent advances of learning informative health indicators, the proposed approaches still often lack generalization to new operating conditions or new fleets, or they can be very sensitive to changing noise levels  \cite{de2023developing}. 
In fact, the lacking robustness of these methods to noise has recently been identified as one of the significant issues of the health indicator construction methods based on machine learning \cite{zhou2022construction}. In this paper, we refer to factors that cause variations in the data (environmental conditions, operational conditions, changing fleets or machines or noise) but are not related to the health condition of the asset as \textit{non-informative factors}.
One of the essential requirements for robust health indicators is to be robust to these \textit{non-informative factors}.
Summarizing, the challenge of constructing {\color{black}versatile} trendable representations of system’s health state (i.e. health indicators) in an unsupervised manner remains unresolved \cite{garmaev2023deep}.

% what are we doing?
In this work, we propose to learn a {\color{black}versatile}, robust and informative health indicator based on contrastive features  that are, on the one hand, (1) very sensitive to slight degradation changes (informative) while, on the other hand, (2) not sensitive towards noise, changes in the operating conditions or changes in the monitored fleet or machine (reliable) and (3) {\color{black}neither} application nor task specific but rather {\color{black}versatile with respect to the application} to condition monitoring data that is either continuously measured or repetitively in distinct time intervals. 
To learn such a robust feature representation, we use contrastive learning {\color{black}respectively} the triplet loss. In absence of labels, we use \textit{operation time} as a proxy for the state of degradation and allow for some self-supervision. On the basis of the resulting feature representation, a health indicator is constructed by measuring the distance to the decision boundary of a One-Class Support Vector Machine (OC-SVM). We evaluate the performance of the proposed method on two datasets {\color{black} with very different characteristics}. First, we conduct experiments on a milling machine dataset for wear assessment for which the ground truth wear condition was continuously monitored on some machines and thus, allows for direct evaluation of the health indicator over time. Still, we train the proposed health indicator in a completely unsupervised way and only use a labeled validation dataset for calibration and evaluation.  Second, we conduct experiments on a real CM dataset of railway wheels from in-operation trains. The {\color{black}wheels are} monitored with Way-Side-Monitoring (WSM) systems{\color{black}, that are able to monitor only parts of the wheel  i.e.\ only parts of the wheel are observed in each measurement}. %As this task is a real CM task of operating trains, 
For this real CM task no ground truth information on the degradation state during operation is available, however, it is assumed that no fault data is represented in the training dataset. In absence of ground truth information, we evaluate our proposed health indicator on the task of fault detection (shelling and crack defects) and  monitor the evolution of the fault condition over time. 
Our results demonstrate that our proposed health indicator  correlates stronger with the real wear of milling machines and improves the performance on the fault detection task regarding the railway wheels as compared to state-of-the-art methods. Further, our experiments are conducted partially on real CM data of real assets in operation and the proposed method is robust to variations in the operating conditions.

The remainder of the paper is organized as follows: in \secref{sec:RW}, the related work is reviewed, followed by the introduction to the case studies in \secref{sec:CaseStudy}. The proposed methodology is introduced in \secref{sec:Methodology}, the performed experiments are detailed in \secref{sec:ExpSetUP} and the results are reported in \secref{sec:Results}. The findings are discussed in \secref{sec:Discussion} and conclusions are drawn in \secref{sec:Conclusion}.

\section{Related Work} \label{sec:RW}

\textbf{Health indicators} are derived from the CM data to describe or quantify the health condition of an industrial asset. 
On the basis of robust health indicators, incipient faults have been detected for e.g.\ wind turbines \cite{chen2020anomaly}, the assest's wear has been assessed for e.g.\ bearings \cite{wang2018theoretical} or the remaining useful life (RUL) has been predicted for e.g.\ an aircraft system \cite{de2023developing}.
While health indicators can be used as inputs for RUL prediction \cite{garmaev2023deep, song2017integration}, which typically requires a sufficient number of time-to-failure trajectories, health indicators can also be applied in cases where labeled training data is absent and used to monitor the evolution of the health condition. Trained in an unsupervised manner, they find application in more realistic scenarios of CM, where labels are often not available. 

As mentioned above, statistical parameters as well as the preprocessing techniques for health indicator construction usually require expert or domain knowledge and are very application specific (see \secref{sec:Introduction}). 
In contrast, machine learning methods are often less application specific and require less domain knowledge.
\citet{chen2022deep}, for example, substituted some domain knowledge by formalizing desired properties of a health indicator as objective functions when learning the health indicator model based on extracted features for the task of RUL prediction.
Most of these proposed machine learning approaches are signal reconstruction-based, where an autoencoder is trained to reconstruct data from the normal system behaviour i.e.\ data that is recorded under healthy conditions \cite{fink2020potential}. 
The health indicator is typically interpreted to represent the distance to the health condition.
For the reconstruction model, different types of neural networks have been proposed: multi-layer perceptron (MLP) models \cite{lu2014intelligent, chen2020anomaly}, convolutional neural networks (CNNs) \cite{guo2018machinery}, 
long short-term memory (LSTM) networks \cite{ye2021health} or Gated Recurrent Unit networks \cite{ni2022data}. 
The health indicator {\color{black}respectively} the distance to the healthy class is then either calculated based on the reconstruction directly \cite{malhotra2016multi, ye2021health} or in the feature space \cite{fu2021novel, zhai2021enabling} or both \cite{liu2020complex}. 
% Input space 
For example, the reconstruction error of a fully connected (MLP) stacked denoising autoencoders has been proposed based on multivariable monitoring data of wind turbines \cite{chen2020anomaly}. The proposed method was evaluated on the task of fault detection. 
\citet{ye2021health} used a multivariate gaussian distribution to construct a health indicator based on the reconstruction error of the proposed LSTM network {\color{black} to quantify} the health state of a  turbofan engine. 
% Feature space
The feature space was considered for health indicator construction by \citet{gonzalez2022health}, where the reconstruction error was extended to the hidden spaces of the autoencoder for training. The health indicator is then  calculated as the reconstruction error in the latent space by passing the reconstructed sample through the encoder model. 
A combination of a signal reconstruction and an OCC-based approach is proposed by \citet{michau2020feature}, where a one-class classifier is stacked onto the features of a hierarchical extreme learning machines (HELM) to monitor the distance {\color{black} from} the test data to the training {\color{black} data which corresponds to the} healthy class.  

Autoencoders are typically trained to reconstruct the data they have seen during training. Any variation in the data could increase the reconstruction error or the distance in the feature space. However, this is only desirable if the change in the data is caused by a change in the health condition, not if the change is caused by noise, changing operating conditions or a new fleet being monitored. The sensitivity to noise of these health indicator methods based on machine learning techniques has been recently identified as one of the key challenges in the field \cite{zhou2022construction}.

Only a few works have addressed the varying operating conditions in the context of health indicators, for example, by providing the operating conditions as an input to the autoencoder \cite{de2023developing}.  \citet{michau2021unsupervised} used adversarial deep learning to transfer health indicators from one operating condition to another.
However, both approaches may lack robustness when faced with operating conditions that are either not  tracktable or have not been encountered during training.

\textbf{Contrastive feature learning}, unlike feature learning based on autoencoders, has demonstrated  robustness against changing operating conditions \citep{rombach2021contrastive}, making it  a promising feature learning paradigm in the context of  non-stationary operating or environmental conditions. The learning objective of the encoder model in contrastive learning is to group data with similar semantic meaning closely in the feature space  while separating  dissimilar data widely  \citep{chopra2005learning}. In supervised settings, the semantic similarity is typically determined by the sample's label \citep{rombach2021contrastive, schroff2015facenet}. 
However, in real condition monitoring (CM) datasets, neither full supervision nor complete  representation of all possible data classes (such as different fault types) is available \citep{fink2020potential}.  Similar challenges have been encountered in other application fields, such as computer vision, where semi- or unsupervised implementations of contrastive feature learning have been proposed \cite{chen2020simple}.
\citet{chen2020simple}, for example, proposed to use augmented versions of the anchor sample as positive samples and treated the remaining samples within a batch as negative samples. It is, however, not straightforward to find appropriate data augmentations for time series CM data.  A proper augmentation would require knowledge on the system itself. Further, since faults occur rarely in operating industrial assets, it usually unknown how faults impact the data. An inproper choice of data augmentation might counterproductive as we may create data augmentations that may resemble faulty data.
For time series data classification, \citet{franceschi2019unsupervised} proposed time-based negative sampling for unsupervised contrastive feature learning. This approach assumes that a randomly picked sample  from a distant point in time would be highly dissimilar to the currently observed sample. This assumption for negative sampling may not hold for time series with e.g. seasonalities (if the sample would coincide with the same periodicity). \citet{franceschi2019unsupervised} selected the positive pair by subsampling the from the same measurement. For CM applications, this positive selection criteria might not be the best choice as it corresponds to data recorded under the same operating and environmental conditions and thus, not result in robustness towards these conditions.

While contrastive learning has not yet been applied for health indicator construction, it has been applied for RUL prediction in an semi-supervised setting \cite{kong2023contrastive}, where it is imposed that the difference of the output of the RUL prediction model from to two unlabeled samples needs to correspond to the difference in time at which these samples have been recorded. However, this approach requires some labeled samples (semi-supervised and not unsupervised) and it learns a model to predict a value that changes linearly in time. This is a restrictive assumption for complex systems the health condition of which is typically not evolving linearly.

\section{Case Studies} \label{sec:CaseStudy}
To assess  the performance of the proposed model, we conducted experiments on two different case studies: 1) the PHM2010 benchmark dataset concerning  the wear of milling machines and 2) a real time-series dataset for CM of  railway wheels. %Both datasets are affected by variability caused by factors that are non-informative of health conditions.

\subsection{Milling Dataset}
The data for the  PHM2010 challenge dataset involving  milling machines \cite{li2009fuzzy} was collected using 3-flute ball cutters on a high-speed milling machine. The milling process was conducted  for dry milling of the slope surface along the horizontal direction \cite{li2022intelligent}. 
The condition of the operating milling machine is monitored using  three types  of sensors: (1) a quartz three-channel platform dynamometer to measure the force along the three axes (x,y,z), (2) three piezo accelerometers along the three axes (x, y and z) and (3) an acoustic emission sensor mounted on the workpiece. 
The available CM data includes  force and vibration measurements in three axes (x, y, z) as well as
acoustic emission signals \cite{li2022intelligent}.
An illustration of the setup is provided in \cite{liu2022three}.
The milling machines are operated under the following conditions: the spindle speed and feed rate are 10,400 RPM and 1555 mm/min, respectively; the Y-direction and Z-direction cutting
depth are 0.125 mm and 0.2 mm, respectively.
The dataset contains data from six different machines, all operated for about 315 cycles (c1, c2, c3, c4, c5, c6) with each cycle lasting 4 minutes \cite{liu2022three}.
The sensory data is recorded at a frequency of 50 kHz. The dataset, although partially unlabeled, presumably contains data that is recorded under a similar wear range. 
For three of the six available machines (c1, c4, and c6), measurements of the actual wear are available in the dataset. 
Between two operation cycles, the flank wear of three flutes
was measured using a LEICA MZ12 microscope. This ground truth information on the wear allows for evaluation of the constructed health indicator. 
While the dataset has been used in previous research studies for wear prediction, most of the studies relied on the labels or domain specific knowledge \cite{liu2022three, li2022intelligent}. %, he2022milling
In our experiments, we only consider the accelerometer data for the following reasons: (a) Since the cutting force signal (in x-direction) has been reported as corrupted \cite{he2022milling, li2022intelligent}, we have chosen to exclude  the measurements from this sensor. (b) Acoustic measurements in an working environment can be affected by a multitude of factors. Therefore, we do not consider both sensors as reliable data sources and use solely on accelerometer measurements in this study. 

For training, we use the three unlabeled machines (c2, c3, c5), for validation and calibration we use the machine c4 and for testing, we use the machines c1 and c6. 

\textbf{Pre-processing:}
As the Melspectogram has proven to be suited for high-frequency time series data \cite{michau2022fully}, we compute the logarithmic Melspectogram from the three accelerometer measurements of each cycle. We used a sampling rate of 50000, a  Fast Fourier Transform (FFT) window length of 5000, a number of samples between successive frames set to 5000 and generated  64 Mel bands. To exclude the non-stationary parts of the measurements, we ignore the first and last three entries. 
Later, we resize the melspectograms to a size of (38, 64) using cubic interpolation and normalize the training dataset and the test dataset is scaled using the parameters from the training dataset normalization.

\subsection{Railway Wheel Monitoring}
For the wheel defect detection dataset, the data was collected from Way Side Monitoring (WSM) systems, referred to as Wheel Load Checkpoints (WLCs), that are distributed over the entire infrastructure network in Switzerland. These WLCs are permanently installed in {\color{black} the} railway tracks and equipped with eight strain gauge sensors on each side of the track such that they capture the data from the wheels on both sides of the train. The sensors measure the vertical force at a frequency of $10 kHz$ when a train passes by, providing information on a part of the wheel's circumference{\color{black}. Since wheel defects are local and do not affect the entire wheel circumference, the wheels are only partially observed} (see \secref{sec:partial}). {\color{black}As mentioned in \secref{sec:Introduction}, in partially observable systems, monotonicity of the health indicator cannot be requested since some measurements might not have captured the defect and correspond to healthy measurements.} The exact setup is described in \citet{krummenacher2017wheel}. Trains pass a WLC approximately five times a day. However, the exact frequency  can vary  significantly depending on the  individual composition's travel plan.  Aside  from changes in the wheels' health conditions, data variability can be caused by 
\textit{non-informative factors} such as different measurement locations and changing environmental conditions. In this study, we focus on monitoring a fleet of passenger trains, which are less prone to wheel flats but more susceptible to other types of defects. We extract all signals for each wheel from each of the eight sensors and concatenate these signals into one measurement. %This means that each sensor measurement covers only specific  parts of the wheel, and the length of each signal depends on the train's speed while passing. 

\begin{figure*}[!t]
\centering
\subfloat[Process flow]{
    \includegraphics[width=3.9in]{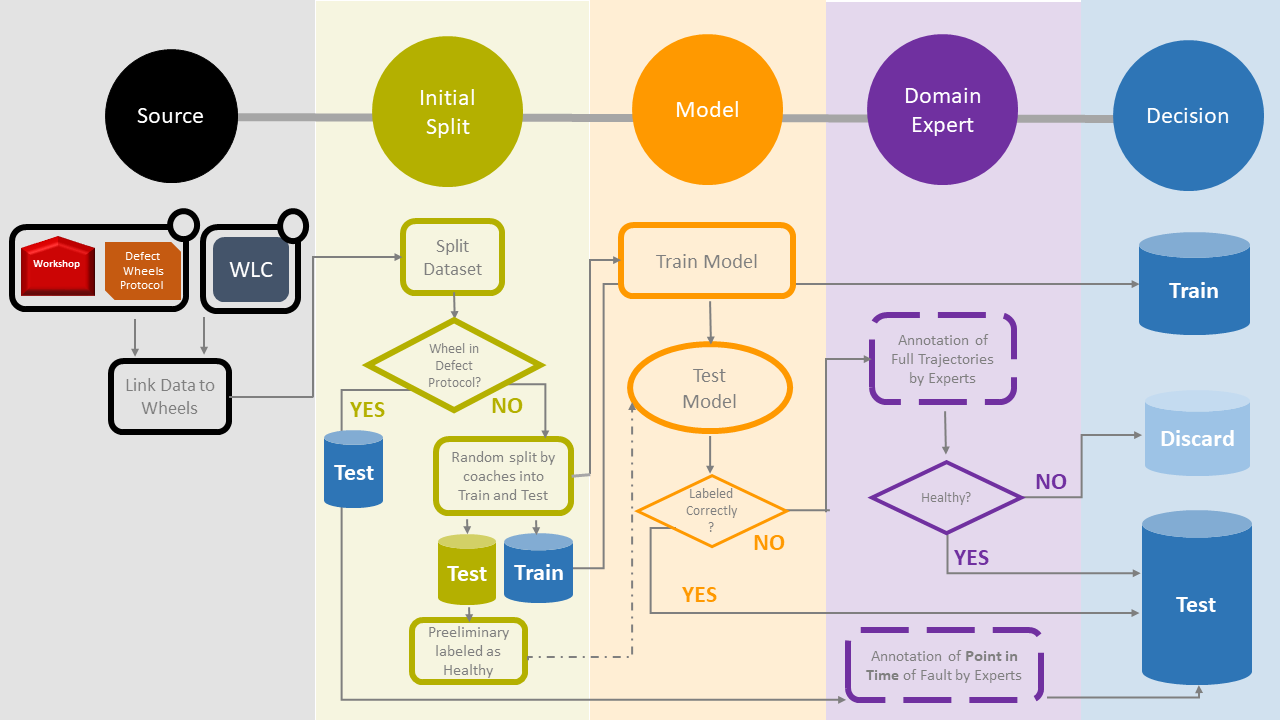} \label{fig:process}}
\\
\subfloat[Point in time annotation]{ \includegraphics[width=3.9in]{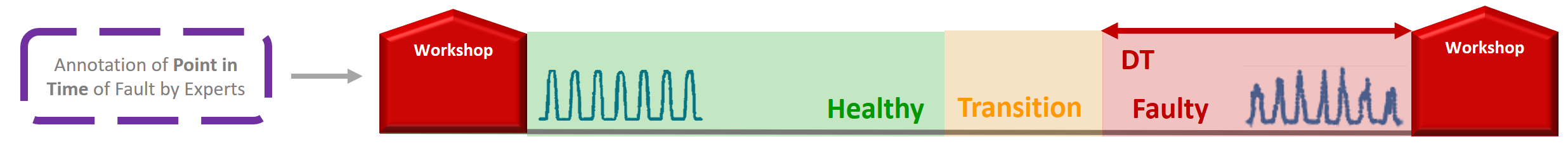}\label{fig:annotation}}
\caption{Data acquisition for the railway  wheel dataset: First in \figref{fig:process}, all data sources are linked to individual wheels ('Source'), resulting in a first data split into train and test ('Initial Split'). For the defective wheels, the time of defect initiation is provided by domain experts, as shown in \figref{fig:annotation}. The preliminary healthy label of the wheels in the test dataset is challenged by fault detection models and evaluated by domain expert feedback.}
\label{fig:process2}
\end{figure*}

For training, 16 coaches (from three different trains) were considered. %However, since coaches with reported wheel defects  were excluded, data from 16 coaches were ultimately used. Consequently, 
No defects were documented for any of the measurements in the training dataset during workshop visits. %, i.e.\ the corresponding wheels can be assumed to be healthy.
However, the supervision process for railway wheels in the workshop visits is susceptible  to errors. It is inherently difficult to detect faults if they are not impacting the operation substantially and visual inspection is the only choice for detection. Cracks, for example, are difficult to detect with visual inspection if the wheels are not very clean.
%due to differing definitions or perceptions of defects, as well as reporting errors. 
Therefore, some measurements in the dataset might correspond to wheels with defects that were not detected by the visual inspection in the workshops. For this work, however, we consider the training dataset as healthy. Because there is  assumed to be an absence of fault data and a lack of supervision in the training dataset, traditional supervised learning as proposed in other studies  (e.g. \citep{krummenacher2017wheel}) cannot be applied.

\begin{figure}[h]
\centering
\subfloat[Crack]{
    \includegraphics[width=0.4\linewidth,angle=90]{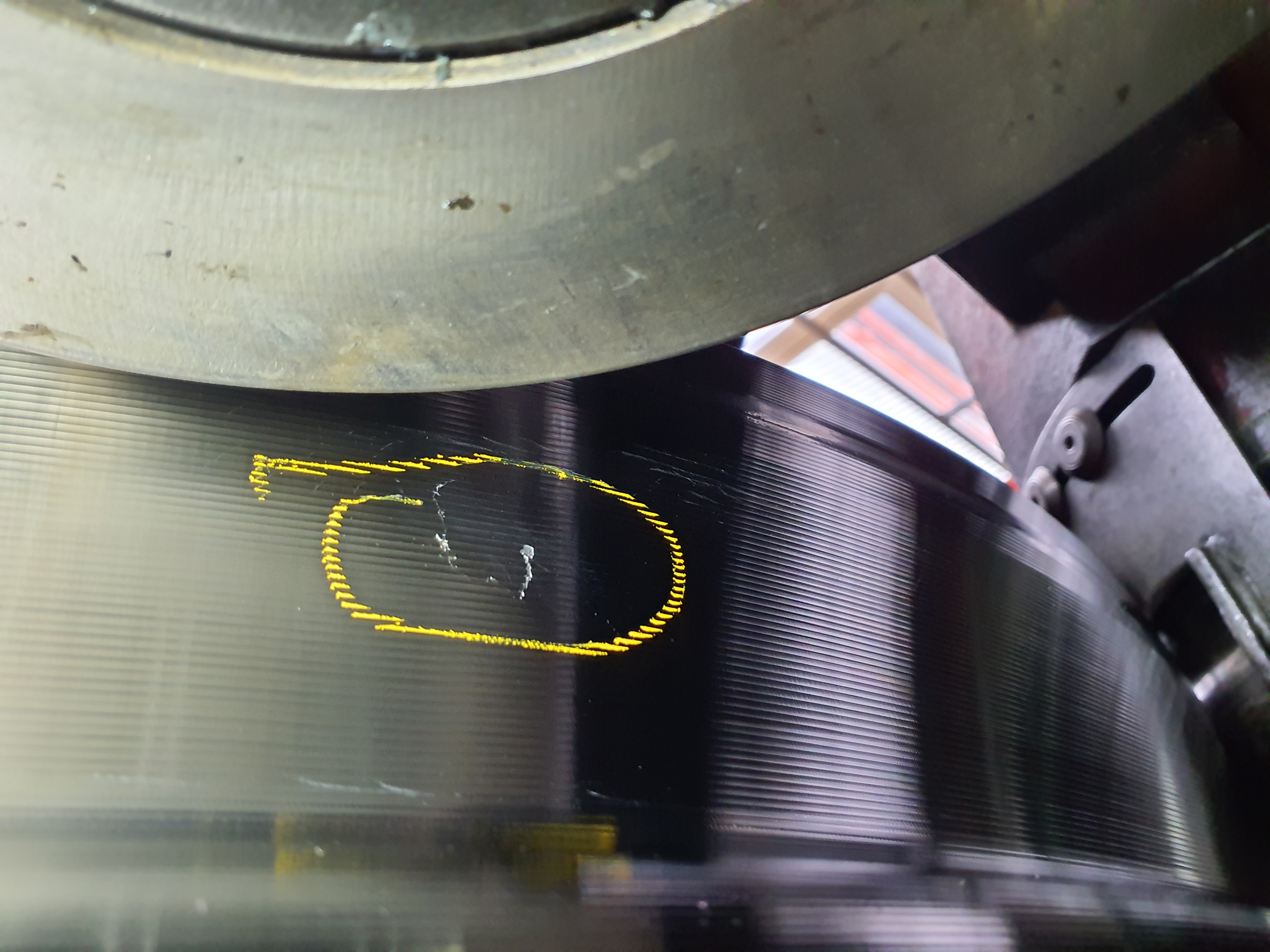}
     \label{fig:crack}}
\hfil
\subfloat[Shelling]{
    \includegraphics[width=0.4\linewidth,angle=90]{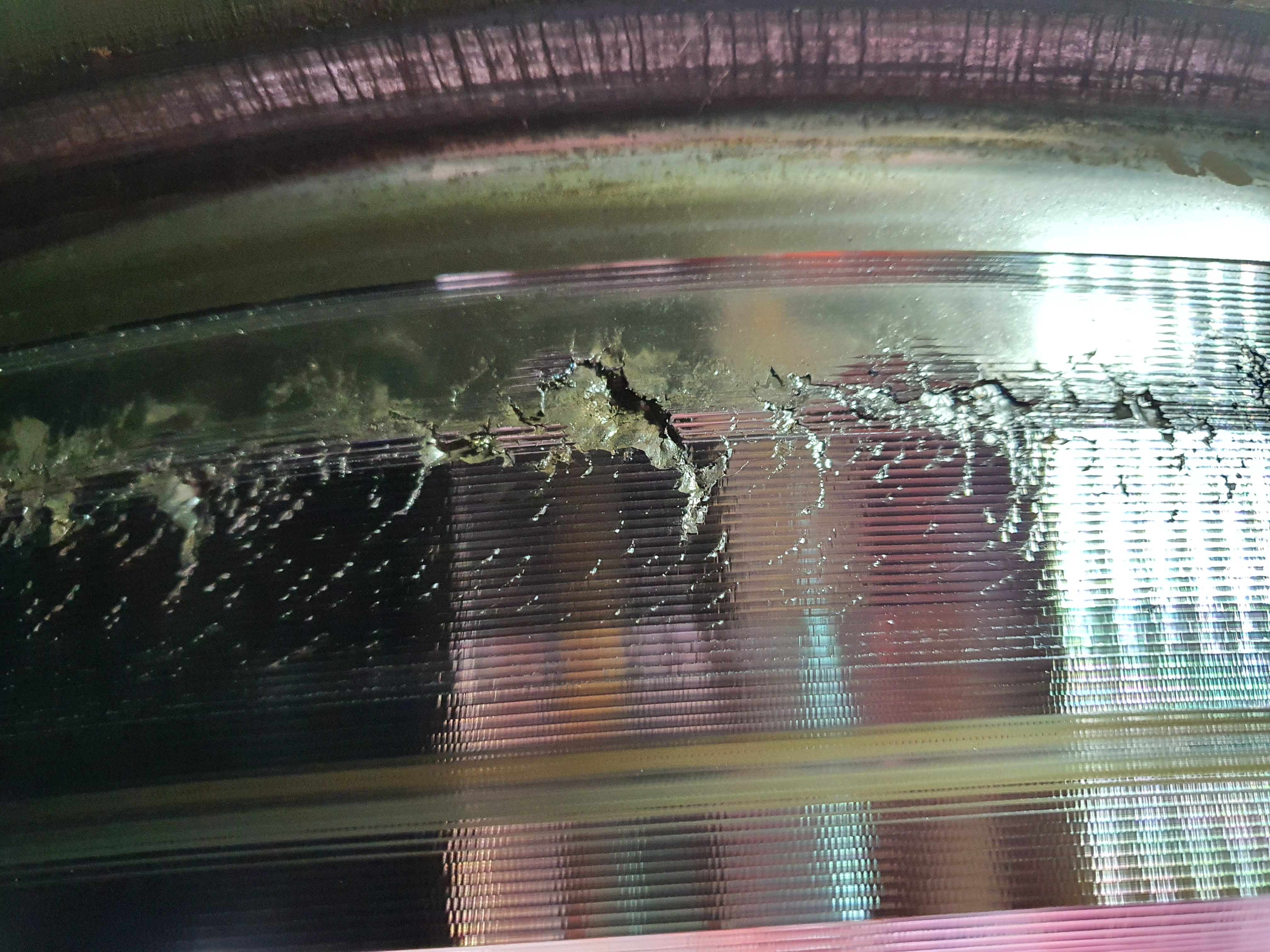}
    \label{fig:shelling}}
\caption{Examples of railway wheel defects}
\label{fig:defects}
\end{figure}

For the test dataset, supervision of the wheel condition is provided during the workshop visit (see 'Sources' in \figref{fig:process}), where the wheels are inspected and re-profiled. A protocol is maintained to report wheels with obvious tread defects. During the one-year monitoring phase, two types of defects  occurred: wheels with cracks (26 wheels) and wheels with shelling defects (53 wheels). Examples of these defect types are shown in \figref{fig:defects}.  An initial test dataset split is performed based on these workshop protocols as illustrated  in \figref{fig:process} (see 'Initial Split'). Due to the difficulties in detecting defective wheels by visual inspection in the workshop, we additionally verified the health status of healthy wheels in the test dataset  that were continuously mislabeled by data-driven models (see 'Model' in \figref{fig:process}) with domain experts and only kept wheels that were confirmed as being healthy (see 'Domain Expert' in \figref{fig:process}). Finally,  the test dataset comprises all data from defective wheels (79 wheels in total) and a randomly selected  subset of healthy wheels (16 wheels). 
%In general, it is difficult to obtain a ground truth health condition of the wheels. On the one hand, wheels are not inspected between workshop visits. Therefore, we do not have any ground truth information on defect initiation. Instead, we rely on the labeling provided by domain experts who investigated the corresponding wheel data as shown in \figref{fig:process} ('Domain Expert') and \figref{fig:annotation}. 
For our study, it is not only important to know which wheels were defect when they visited the workshop, we also need to know when the defect has happened. Since this information is not available, we asked domain experts to label the test dataset as follows: 
\begin{itemize}
    \item The green zone corresponds to the timespan in which the fault has not yet manifested in the data. 
    \item The orange zone corresponds to the transition phase in which the fault  starts to manifest itself in the data. The domain experts were asked to mark the first possible point in time they would be able to detect the fault based on the measurement data.
    \item The time period after the orange zone is marked as the red zone.
\end{itemize}
 %The red zone corresponds to the time period in which   the fault can be detected by domain experts based on the data assessment.
%Wheels that have not been reported as having defects might nevertheless be defective. Different degrees of attention or detail of reporting by the inspectors, as well as non-obvious defect types, can quickly lead to overlooked defects - also in the test dataset.  Since the exact condition of the presumably healthy wheels in the test dataset is not known, the measurements from healthy wheels that are labeled as faults by the fault detection model (see 'Model' in \figref{fig:process}) are discussed with domain experts (see 'Domain Expert' in \figref{fig:process}). Only those data samples of the wheels that are labeled as healthy by the domain experts are added to the test dataset (see 'Decision' in \figref{fig:process}). 

\textbf{Pre-Processing:} First, we resample  the concatenated signals using  linear interpolation to standardize the length to 1024 samples,  accounting for variations in train speed. Next, we normalize each recorded signal to account  for differences in train loads.

\subsubsection{Partial Observable Railway Wheels}
\label{sec:partial}
\begin{figure*}[h]
\centering
\includegraphics[width=0.5\linewidth]{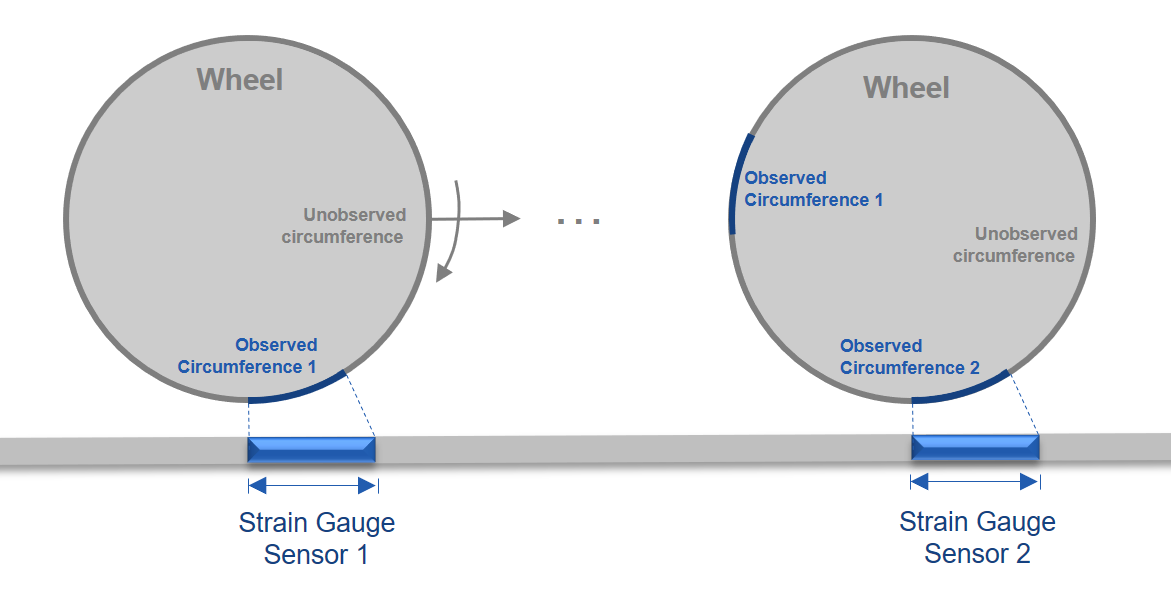}
\caption{Illustration of the partial observation of railway wheels provided by a strain gauge sensor in a WSM measurement site.} \label{fig:partial}
\end{figure*}

Since the measurements of a WLC cover only parts of the wheel's circumference, reliable health monitoring is challenging as local defects might fall outside of the measurements and data samples from defective wheels might not show any signs of faults in the data. %For example, WLC measurement sites with strain gauge sensors in Switzerland (see \secref{sec:CaseStudy}), for example, are only capable of covering parts of the railway wheel i.e. the wheel's condition is only observed partially by the measurement system. 
This is illustrated in \figref{fig:partial}. % whereas a lower limit of the observed region on the wheel per sensor is 28 cm. 
%A wheel defect  might not be visible in the recorded CM data and therefore, the defect is not detectable. 
%A lower limit of this observed region per sensor is 28 cm. 
%Since individual measurements monitor only parts of the asset, data samples from defective wheels might not show any signs of faults in the data. 
Therefore, multiple consecutive measurements are required to decide on the asset's condition since this increases the probability that a defect is observed by the CM data. %Furthermore, we aim to desensitise the fault detection model to variations caused by domain shifts e.g.\ due to measurement site calibration. 
We consider 5 consecutive measurements to decide on the asset's condition.
%To account for that, we choose to model the probability of a fault being sufficiently represented in the data with a binomial distribution, whereby we assume that the defect should be represented in at least $K$ out of $N$ measurements. 
such that monitoring within a day is possible. 

\begin{figure*}[h]
\centering
\includegraphics[width=0.7\linewidth]{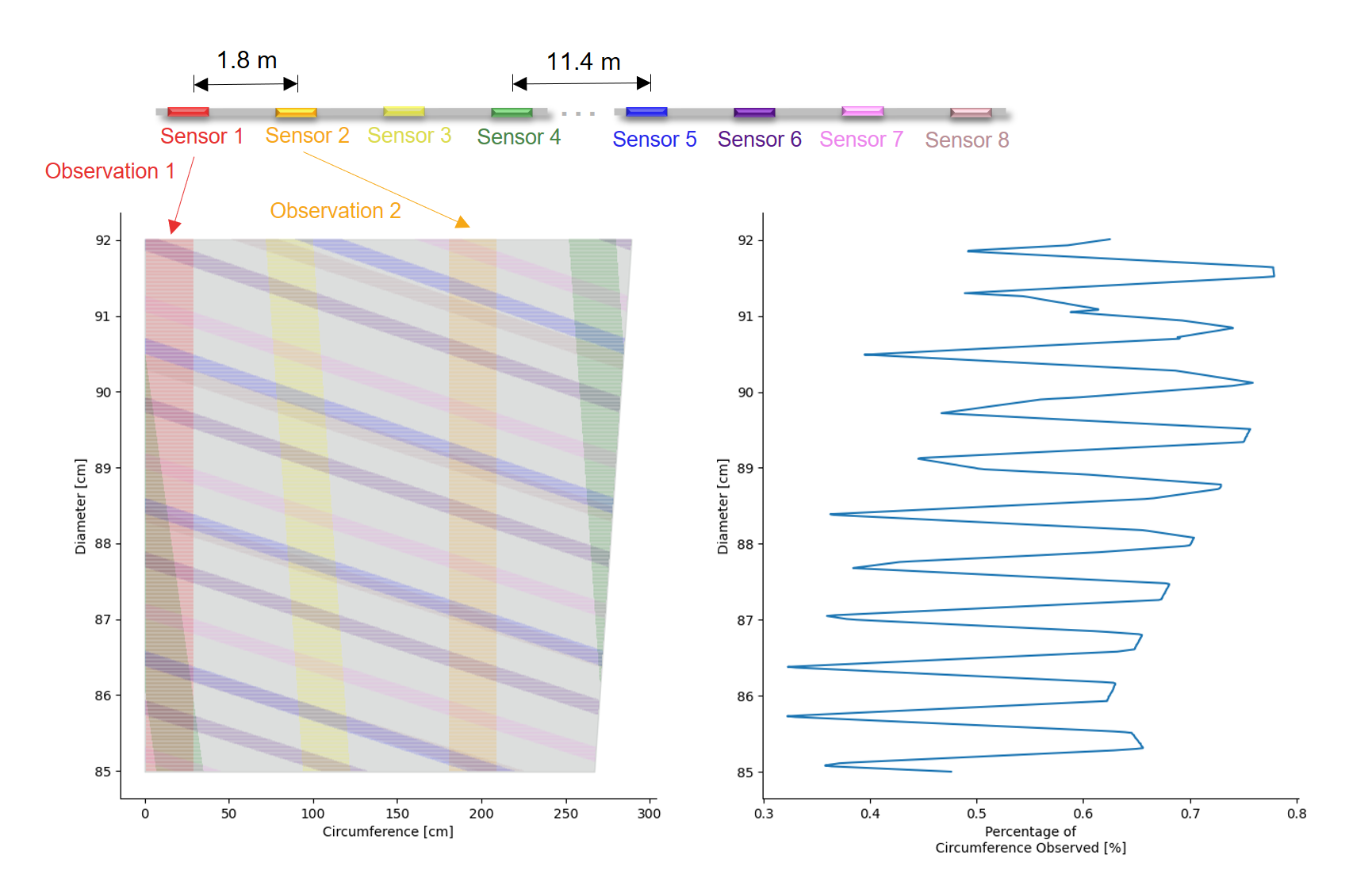}
\caption{Wheel circumference regions  monitored by the eight strain gauge sensors (blue regions) in dependency of different diameters compared to the entire wheel circumference.} \label{fig:wheelcircumference}
\end{figure*}

The probability of a fault being represented in an individual measurement from the WSM site corresponds to the percentage of the wheel circumference that is covered by the strain gauge sensors, which depends on the current diameter of the railway wheel. This is shown in \figref{fig:wheelcircumference} where the colored regions correspond to the parts on the circumference being monitored by the eight different sensors. The figure was produced with the lower limit of possible circumference coverage of the measurement site (28 cm of the circumference being observed by an individual sensor) and provides an lower limit of the probability of a defect being represented in a measurement. Moreover, wheel diameters are used that are within the specifications of the monitored fleet. 
To account for this partial observability, in this study, a wheel is labeled defective if the median of 5 consecutive health indices is above the defined threshold. 

\section{Methodology} \label{sec:Methodology}
In this research, we propose, a two-step process for learning {\color{black}versatile}, robust and informative health indicators. The process and its details are described below. 

\subsection{Proposed Framework} \label{sec:framework}
\begin{figure*}[h]
\centering
\includegraphics[width=0.6\linewidth]{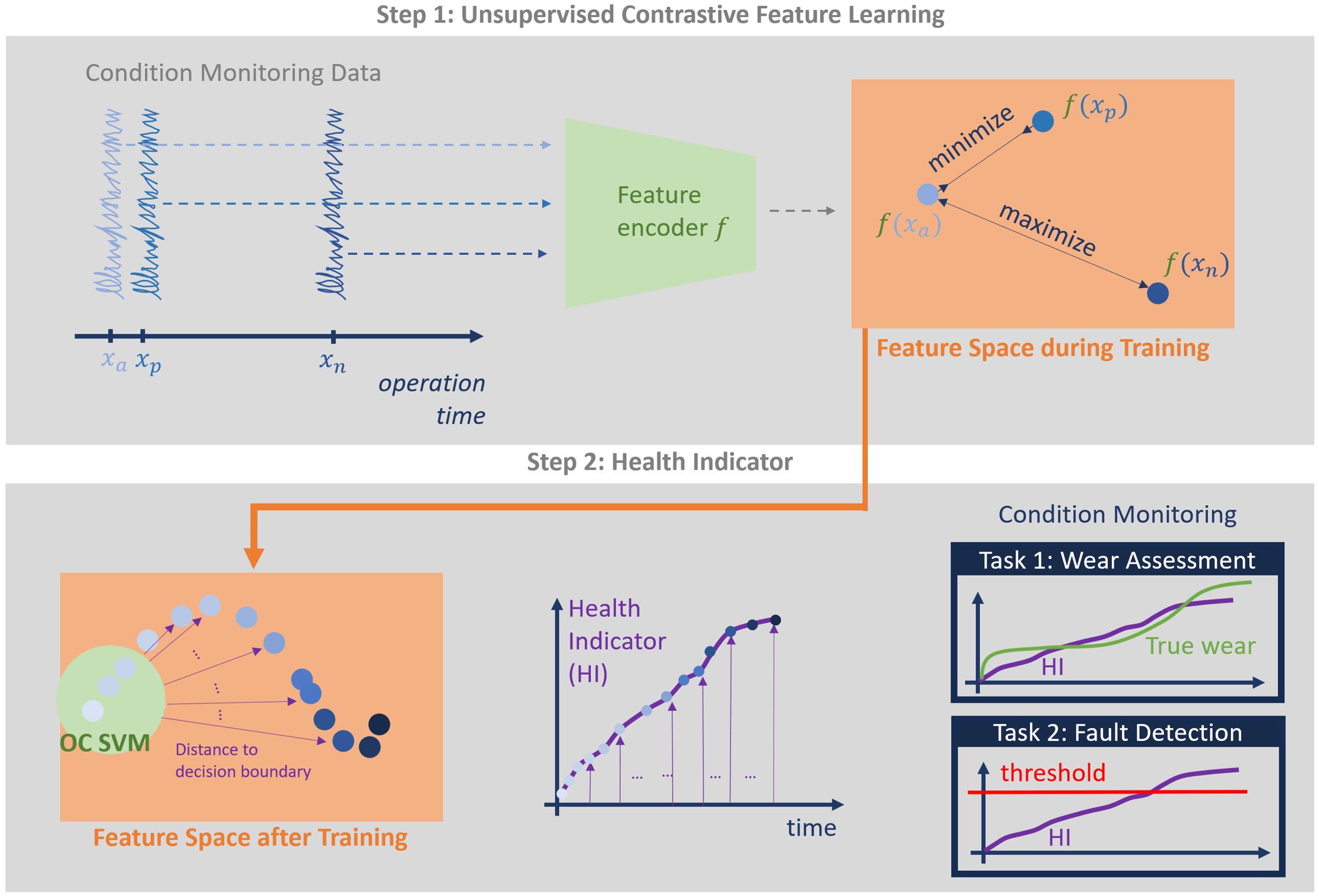}
\caption{Illustration of the proposed framework: In a first step, an encoder model $f$ is trained on with the triplet loss function $L(x_a, x_p, x_n)$, whereby the selection of the data triplets $x_a, x_p$ and $x_n$ is based on the \textit{operation time}. In a second step, the trained feature space is exploited to construct the health indicator for CM: wear assessment (upper right) and fault detection (lower right)} \label{fig:framework}
\label{fig:method}
\end{figure*}

%In the first step, we propose to utilize contrastive feature learning for learning a compact representation of the CM data that is, on the one hand, invariant to data variations caused by non-informative factors and, on the other hand, sensitive to  changing health conditions. We showcase how this can be achieved for tasks with different degrees of data and label availability within the railway system. Based on the learned feature representation, in a second step, the condition of the assets is monitored by implementing a fault detection or diagnostic solution. The entire methodology in its supervised and unsupervised application is illustrated in \figref{fig:method} 
The proposed methodology comprises two steps: 1) In the first step, we propose unsupervised contrastive feature learning for learning a compact feature representation of the CM data by using the \textit{operation time} as a proxy for the degradation state of the asset (see  \secref{sec:contrastive_meth}). A simplified illustration of the contrastive selection process is shown in \figref{fig:method}.  2) In the second step, a health indicator is constructed based on the learned feature representation by measuring the distance to the decision boundary of a{\color{black}n} OC-SVM (see \figref{fig:method}). Based on this health indicator, the condition of the assets is monitored over time. %To showcase the versatility of the proposed framework, we assess the wear of assets on one case study and we detect faults on another case study  ( see \secref{sec:hm_meth}). 
The entire methodology for its different CM tasks (unsupervised wear assessment and fault detection) is illustrated in \figref{fig:method}.

\subsubsection{Unsupervised Contrastive Feature Learning}
\label{sec:contrastive_meth}
The core of the proposed methodology is the contrastive loss function with which we train an encoder model to impose invariance to \textit{non-informative factors} and sensitivity to changing health conditions.
%a two-step process for railway fault detection and diagnostics tasks: In a first step, a contrastive feature representation is learned (either supervised or unsupervised) and in a second step, the detection and diagnostics model is trained based on the learned features (either classification or OCC).
More concretely, the \textbf{semi-hard triplet loss} is used as defined in \equref{eq:contrastive}, whereby  $x_a$ is the anchor sample, $x_p$ the positive sample (that shares the semantic meaning {\color{black}respectively the same} health condition with the anchor), and $x_n$ is the negative sample with a different semantic meaning {\color{black}i.e.\ with a different} health condition \citep{schroff2015facenet}, $f(.)$ is the encoded sample, $||.||$ is a distance metric, and $\epsilon$ is a margin parameter.  By minimizing the distance between the anchor and positive sample pair, the invariance to non-informative variations is imposed and by increasing the distance between the anchor and negative pair, the sensitivity to faults is increased. 

\begin{equation}\label{eq:contrastive}
\begin{aligned}
 L(x_a, x_p, x_n) & =  \sum_{x_a \in X} max(0,\sum_{x_p \in P}(|| f(x_a) -  f(x_p)||) \\ & -  || f(x_a) - f(x_n)|| + \epsilon) 
\end{aligned}
\end{equation}

In absence of labeled data, the data triplets i.e.\ samples that share the same health condition (for positive pairs) or have dissimilar health conditions (for negative pairs) is not possible. 
To allow for unsupervised selection of data triplets ($x_a$, $x_p$ and $x_n$), we exploit the fact that (a) the condition of industrial assets is typically monitored continuously or in discrete time periods and (b) the health condition of each asset typically changes similarly over time due to normal degradation, if operated in a similar manner.
Thus, we use the \textit{operation time} as a proxy for the health condition. The \textit{operation time} is defined as either the time that has passed since a machine was taken in operation or the time that has passed since the asset was last maintained. %The selection of the data triplets is similar to the one proposed by \citet{franceschi2019unsupervised} with the difference, that we consider the operational context for the triplet selection selection in order to (a) achieve invariance to operating conditions, measurement sites or a change of machines that is being monitored while (b) enforcing sensitivity to small degradation changes that might not be as prominent in the data as the other \textit{non-informative factors}. 

\textbf{Unsupervised choice of positive samples:} Measurements that are taken close in \textit{operation time} are considered to have the same health condition but are recorded under different \textit{non-informative factors} (other machines, other measurement sites, different loads etc.\ ). By considering these measurements as the positive pairs we impose invariance to the \textit{non-informative factors}. In this study, we use a fixed timespan in which measurements are considered as positives.
To emphasize on the invariance within the same health condition across different machines, operating conditions or measurement sites, the average distance of multiple positive samples within a batch of data is considered. 
If no validation dataset is available to tune the encoder training, all positive samples within the defined timespan are considered in the loss function per anchor sample. 
If a validation dataset is available, we additionally allow for some self-supervised selection of n positive samples, whereby the n closest samples are considered in the loss function. 
This self-supervised selection of the positive samples is useful if the machine's health condition evolves differently in time and therefore the proxy of \textit{time of operation} is a noisy measure for the asset's health condition.

\textbf{Unsupervised choice of negative samples:} Measurements in between which a sufficient amount of  \textit{operation time} has passed, are considered to have different health condition{\color{black}s} (negative pair).
To  enable smooth learning, a semi-hard negative sample is chosen as defined in \cite{schroff2015facenet}. A semi-hard negative sample is the closest negative sample outside of the sphere in the feature space of all samples that are recorded within a timespan that defines the positive samples. 

\subsubsection{Health Indicator}
\label{sec:hm_meth}
After training the encoder model as described above, the learned feature representation is used for the health indicator construction.
%Since the feature space already is trained to filter any \textit{non-informative factors} and be sensitive to degradation, the health indicator contruction method is kept not complex. 
We train an OC-SVM model on the features of the healthy data in the training dataset (where the \textit{operation time} is short).  
The OC-SVM builds a hypersphere around the healthy data that represents a decision boundary to distinguish data that was recorded under the same health conditions as the training data (healthy data with a short \textit{operation time}) and the data that was recorded under different health conditions.
As the feature space is constructed such that similar health conditions are in close proximity in the feature space and dissimilar health conditions are far apart, the health indicator is constructed by measuring the euclidean distance to the decision boundary of the OC-SVM in the feature space. 
%The feature space is trained to be semantically feasible in \secref{sec:contrastive_meth}. I.e.\ we aimed to group
%If the feature space is trained to be semantically feasible i.e.\ the encoder model trained in \secref{sec:contrastive_meth} is capable of grouping 
%similar health conditions close to each other, data from slightly different health conditions slightly further apart and data from substantially different health conditions far apart in the feature space. If this is the case, the distance from the healthy class in the feature space can be representative of the severity of a defect. 
The distance of a sample to the decision boundary of the OC-SVM, corresponds to its distance to the healthy training data. We therefore interpret this distance as the health indicator that is representative of degree of degradation or the severity of a defect. The he{\color{black}a}lth indicator construction is illustrated in \figref{fig:method}. %Therefore, the distance in the feature space to the decision boundary of the OC-SVM represents the health index that is used to assess the wear over time and detect anomalies.

A good health indicator should {\color{black}satisfy some} properties (see \secref{sec:Introduction}) to be useful and trustworthy. {\color{black} It should have a smooth and, for fully observable systems also a} monotone evolution. Thus, we apply two post-processing steps: First, we smooth the health indicator by calculating the moving average of consecutive measurements. We do this to account for noisy measurements. Second, for wear assessment, we only report the last highest value of the smoothed indicator.

\subsection{Performance Evaluation}
\label{sec:evaluation}
\textbf{Wear Assessment of Milling Machines:} After scaling the health indicator based on the validation dataset, we calculate and report the correlation and cosine similarity between the constructed health indicator and  the true wear of the asset. Furthermore, for  the trendability of the health indicator we report the Spearman correlation coefficient {\color{black} as suggested in previous work} \cite{de2023developing}, as it is one of the  primary  requirements  for health indicators. For evaluation purposes only, we also scale the health indicator to match the range of the wear values of the considered machine and report the mean squared error between the health indicator and the true wear.     

\textbf{Unsupervised Health Monitoring of Railway Wheels:}
We report the {\color{black}confusion matrix as well as the} balanced detection accuracy of the wheels in the test dataset. Additionally, we evaluate the timing of defect detections (see \figref{fig:annotation}), as it is crucial to detect defects as early as possible. 
For defective wheels, we measure the time interval $dt$ (number of days) between the detection and when the wheel defect manifests itself in the CM data, as indicated by expert labeling (see \figref{fig:annotation}). A negative value ($dt<0$) corresponds to detection in the green zone (see \figref{fig:annotation}), a positive value ($dt>0$) corresponds to detection in the red zone, and $dt=0$ corresponds to a detection in the orange zone.  Because  detections in the red zone can potentially be critical,  and the length of the red zone can vary  substantially between different wheels, we further  assess  them  using  a relative measure $dr$ of the total delay time interval $dt>0$ in relation to the entire time interval $DT$ in which the defect was present during operation (see \figref{fig:annotation}). An early detection (in the green zone) may be favourable for early maintenance planning but could also indicate that the model is too sensitive for real operational use. Therefore, we consider detection in the orange zone as desirable. 

\subsection{Alternative Methods for Comparison}
\label{sec:comparison}
We use HELM as a comparison method. This feature learning method  also enables the  tracking of evolving conditions over time in the form of a health ind{\color{black}icator} and was used in previous case studies with similar setups \citep{michau2020feature}. Additionally, this method is {\color{black} also versatile as it is} suitable not only for degradation assessment but also for fault detection. Unlike the proposed method, HELM's  feature encoder model is trained to reconstruct the signal, leading to a different model architecture choice. %The number of units per layer was determined such that the reconstruction loss was sufficiently decreased. The encoder architecture is a five-layer network with 30 units each and 100 units for the OCC. The remaining parameters are set to default values ($C=1e-5, \lambda=1e-3$).
%The AD detection threshold is set as described in \secref{sec:Methodology}.

For the unsupervised fault detection task on the railway case study, we apply an additional method for comparison that is used currently in operation with a threshold of $1.8$. We compare our results to a statistical measure known as  the dynamic coefficient (\textit{dynCoeff}) -  as described in \equref{eq:db}. This coefficient quantifies  the ratio of the maximum dynamic to the static wheel load within each sensor measurement $x$.

% Currently %It is a general measure of spread within each sensor measurement $x$, that is currently 
%used in operations with a threshold of 1.8.

\begin{equation}\label{eq:db}
\begin{aligned}
dynCoeff & =  \frac{max(x)}{mean(x)}
\end{aligned}
\end{equation} 

{\color{black}Further, for completeness, we also report the results on this case study obtained by using the distance to the decision boundary of the OC-SVM, where we use the exact same settings as in the {\color{black}C}ontrastive model, just based on the preprocessed data instead on the extracted features.}

\section{Experimental Setup}
\label{sec:ExpSetUP}

\subsection{Milling Degradation Assessment}  
A two-layer 2D convolutional model (with ReLu activation) is used with 32 {\color{black}respectively} 64 filters and a kernel size of four in each layer.
Last, a fully connected layer is added with 10 nodes. The architecture is chosen based on the correlation value {\color{black}of the health indicator} to the wear in the validation dataset (c4). 
The model is trained with a batch size of 256 and the Adam optimizer with a learning rate of (0.0002) for 1000 epochs.
{\color{black}The positive timeframe, within which positive samples are selected, is set to be $+/-$ three cycles from the anchor sample; that is, data recorded in any of the three cycles before or after the anchor sample. Within this timeframe, the samples  are} considered to share a similar health condition.
An OC-SVM is applied to the extracted features with a linear kernel function on the first five cycles of each of three milling machines. The health ind{\color{black}icator} is calculated at test time as the distance to the decision boundary of the OC-SVM.

The comparison method HELM is trained using a single layer AE with 100 neurons, and a one-class classifier with 100 neurons. The network architecture was tuned to maximize  the correlation value on the validation data rather than minimizing  the reconstruction error, to ensure a fair comparison. The multiplicative factor for determining the threshold is set to 1.0, and the threshold is determined  based on 99.9\% of the training data, following the same setting as for the {\color{black}C}ontrastive model.  An ensemble of five models was trained and run. We used standard values for HELM, with $C=1e-5$ and $\lambda=1e-3$. HELM, being a reconstruction-based method, relies on a higher  reconstruction error for more degraded conditions. Therefore, for the wear assessment task, we trained the HELM only on data with a short \textit{time of operation} ($<$50 cycles).

\subsection{Health Monitoring Algorithm for Railway Wheels}
We used a five-layer 1D convolutional model with ReLu activation at a degree of 0.1. Each layer consists of 10 filters and a kernel size of 16.
Additionally, a fully connected layer with four nodes was added. This architecture was determined through validation on a dataset comprising $10\%$ of the entire training dataset. We aimed  to keep the feature space dimensionality as small as possible (set to four) to encourage the encoder model to focus solely  on  relevant data variations. The model was trained using the Adam optimizer with default settings for 100 epochs and a batchsize of 32.
To calculate the contrastive loss function, we define positive pairs as data measurements recorded within the same month after the workshop visit. This timeframe is selected  based on domain knowledge. 
We apply an OC-SVM to the extracted features using a Radial Basis Function (RBF) kernel function and set a threshold of 0.88. The choice of a relatively low threshold is influenced by the characteristics of the real data. Firstly, the exact condition of the training dataset is not known. Secondly, individual sensors can be poorly calibrated, leading to anomalies in the training dataset. The health ind{\color{black}icator} is calculated during  test phase  as the distance to the decision boundary of the OC-SVM.
The comparison method HELM is trained using a single layer AE with 30 neurons, and a one-class classifier with 100 neurons. The multiplicative factor for determining the threshold is set to 1.0, and the threshold is established  based on 88\% of the training data, following the same setting as for the {\color{black}C}ontrastive model. An ensemble of five models was trained and executed. Other values were chosen to be the standard values for HELM ($C=1e-5$, $\lambda=1e-3$).

\section{Results} \label{sec:Results}
%The results obtained by the conducted experiments are reported below. The result on the wear assessment of milling machines is reported in \secref{sec:results_milling}, the results on the unsupervised case of railway wheel monitoring is reported in \secref{sec:results_wheel}.

\subsection{Milling Degradation Assessment}
\label{sec:results_milling}
\begin{figure*}[h]
\centering
\subfloat[Health indicator machine 1 (c1)]{
    \includegraphics[width=0.4\linewidth]{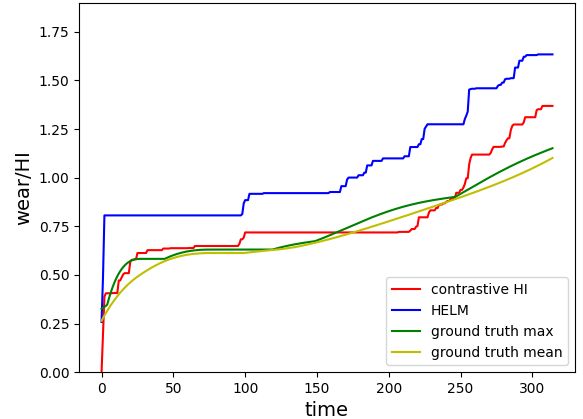}
    %\caption{Health indicator Flute 1 (c1)} 
    \label{fig:hi_f1}}
    \hfil
\subfloat[Health indicator machine 2 (c6)]{
    \includegraphics[width=0.4\linewidth]{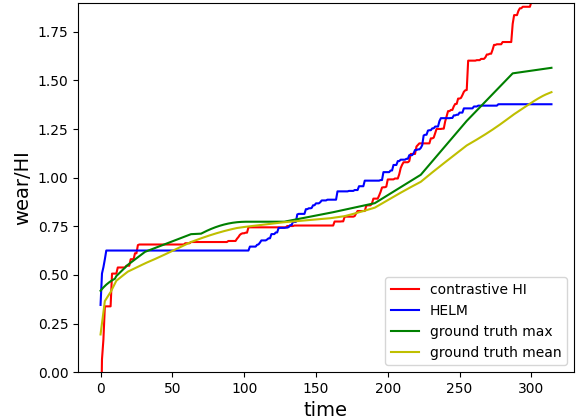}
    %\caption{Health indicator Flute 2 (c6)}
    \label{fig:hi_f2}}
\caption{Health indicator of milling machines. The health indicator based on HELM for the two test flutes are plotted in blue. The health indicator based on the proposed health indicator in this paper for the two test flutes are plotted in red.}
\label{fig:hi_milling}
\end{figure*}

\begin{figure*}[h]
\centering
\subfloat[Machine 1 (c1)]{
    \includegraphics[width=0.4\linewidth]{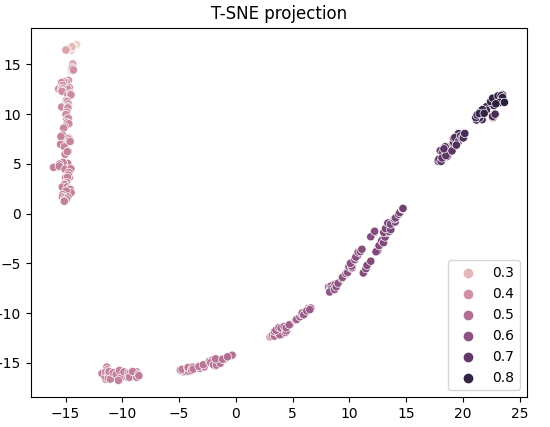}
    %\caption{ Flute 1 (c1)} 
    \label{fig:feature_f1}
}
\hfil
\subfloat[Machine 2 (c6)]{
    \includegraphics[width=0.4\linewidth]{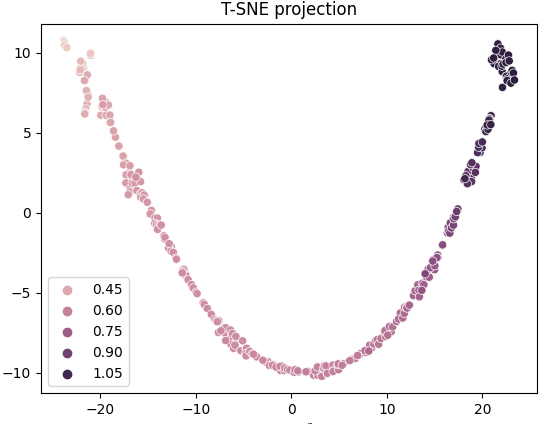}
    %\caption{Flute 2 (c6)}
    \label{fig:feature_f2}
}
\caption{TSNE plot of the contrastive feature space for wear assessment.}
\label{fig:features_wear}
\end{figure*}

The results for the wear assessment of milling machines are presented below.
In \tabref{tab:correlation_value},  we illustrate the relationship between  the proposed health indicator and the {\color{black}maximal} wear of the three flutes of the milling machine. 
%We report both the correlation and the cosine similarity between  the health indicator and the wear of the three flutes. Additionally,  we provide metrics related to individual flutes,  as well as  metrics related to the maximal wear and the mean wear of the flutes. 
\figref{fig:hi_milling} displays the health indicator trajectories of the proposed method and HELM. In \figref{fig:features_wear}, we visualize the feature space of the contrastive features for the two test machine.  All reported results are the best results selected from 10 iterations based on the validation dataset. %In the \secref{sec:appendix} a more finegrained evaluation is available. 

In the context of wear assessment, our proposed health indicator performs  better  when compared to the actual  wear of the assets.   It demonstrates higher correlation (on average +0.03) and  a better trendability  (on average +0.05) in comparison to the state-of-the-art method. Only, the mean-squared error is slightly lower for the HELM indicator (-0.01). However, this metric is scaled based on the real wear values on each machine and is thus, the least expressive metric.  Notably, in the initial phase (cycles $<$ 150), our proposed health indicator closely follows the true wear pattern  compared to the HELM indicator. The transition in dynamic, where the wear begins to increase at a higher rate, is evident  in the trajectories of each health indicator but slightly more accurate for the contrastive health indicator on machine c6. For both machines, the contrastive health indicator shows the same behaviour by overestimating the wear. The HELM health indicator, however, shows different behaviour by once overestimating the wear by a larger margin as the contrastive indicator (c1) and once underestimating it (c6).  

\begin{table*}[h]
\caption{Evaluation of the health indicator for wear assessment of the two milling machines in the test dataset: The correlation (Corr) and the Cosine-Similarity (Cos) between the health indicator and the maximal wear, the Mean-Squared-Error (MSE) between the scaled health indicator and the maximal wear as well as the trendability (Trend) \label{tab:correlation_value}}
\centering
\begin{tabular}{|c | cccc | cccc |}
\hline
&\multicolumn{4}{c|}{Flute 1 (c1)} & \multicolumn{4}{c|}{Flute 2 (c6)} \\ \hline
& Corr & Cos & MSE (scaled) & Trend & Corr & Cos & MSE (scaled) & Trend\\ \hline

%OC-SVM & 0.94 & 0.99 & 0.02 & 0.97 & 0.94 & 0.99 & 0.1 & 1.00 \\[0.2ex]
    
HELM    & 0.92 & 0.97 &  0.01 & 0.92  & 0.95 & 0.99 & 0.02 & 0.95   \\[0.2ex]
Contrastive (ours)    & 0.94 & 0.99 & 0.01 & 0.98 & 0.99 & 0.99 & 0.04 & 1.00 \\[0.2ex]
\hline
\end{tabular}
\end{table*}

\subsection{Railway Wheel Monitoring - Fault Detection}
\label{sec:results_wheel}
The results from the railway wheel case study are presented based on the extracted health indicator and the decision rule for partial observable measurements. First, the fault detection results are displayed and second, the detection time is evaluated (see \secref{sec:evaluation}). 
In \tabref{tab:Wheels_AD},  we present the results of the anomaly detection task. The \textit{dynCoeff} model appears to be the least sensitive model to faults, correctly identifying all healthy wheels but detecting only four defective wheels. {\color{black}The OC-SVM model shows considerably higher sensitivity to the defects, detecting the most defective wheels (67). It, however, also has the highest rate at falsely detecting healthy wheels as faulty, resulting in the highest false alarm rate.}  In contrast, both feature learning methods {\color{black} detect} 63 out of 79 faults but also mislabel only one of the healthy wheels as faulty. Additionally, we report the results of an ensemble composed of HELM and the Contrastive model. The decision rule for the ensemble is as follows: Any wheel detected as having a defect by either of the models (HELM or Contrastive) is labeled as defective. {\color{black} This model results in the highest balanced accuracy of $88.7\%$.}

\begin{table*}[h]
\tiny
\caption{Fault detection results on the railway wheel dataset. \label{tab:Wheels_AD}}
\centering
\begin{tabular}{|cc | cc |cc|cc|cc|cc|}
\hline
&&\multicolumn{10}{c|}{Predicted} \\ \hline
     &
    &\multicolumn{2}{c|}{\textit{dynCoeff}} &\multicolumn{2}{c|}{OC-SVM} &\multicolumn{2}{c|}{HELM}  &\multicolumn{2}{c|}{Contrastive} &\multicolumn{2}{c|}{HELM + Contrastive} \\ \cline{3-12}
     & 
     & 
    \multicolumn{1}{c}{Defect} & 
    \multicolumn{1}{c|}{Healthy} &
    
    \multicolumn{1}{c}{Defect} & 
    \multicolumn{1}{c|}{Healthy} &
    
    \multicolumn{1}{c}{Defect} & 
    \multicolumn{1}{c|}{Healthy} &

    \multicolumn{1}{c}{Defect} & 
    \multicolumn{1}{c|}{Healthy} &
    
    \multicolumn{1}{c}{Defect} & 
    \multicolumn{1}{c|}{Healthy} \\ 
    \hline
    \multirow[c|]{2}{*}{\rotatebox[origin=tr]{90}{Actual}}
    & \multirow[l]{1}{*}{Defect} & 4 & 75 & 67 & 12 &  63 & 16  & 63 & 16  & 71 & 8 \\ %[0.2ex]
    & \multirow[l|]{1}{*}{Healthy}  & 0   & 16 & 3 & 13  & 1 & 15  & 1 & 15 & 2 & 14\\ [0.2ex]
    \cline{2-12}
    & \multirow[l]{1}{*}{Balanced Accuracy}  & \multicolumn{2}{c|}{52.5\%}&  \multicolumn{2}{c|}{83.0\%}  &  \multicolumn{2}{c|}{86.7\%}  & \multicolumn{2}{c|}{86.7\%} & \multicolumn{2}{c|}{\textbf{88.7\%}} \\
\hline
\end{tabular}\\[10pt]
\end{table*}

\begin{figure*}[h]
\centering
\subfloat[Shelling defect]{
    \includegraphics[width=0.4\linewidth]{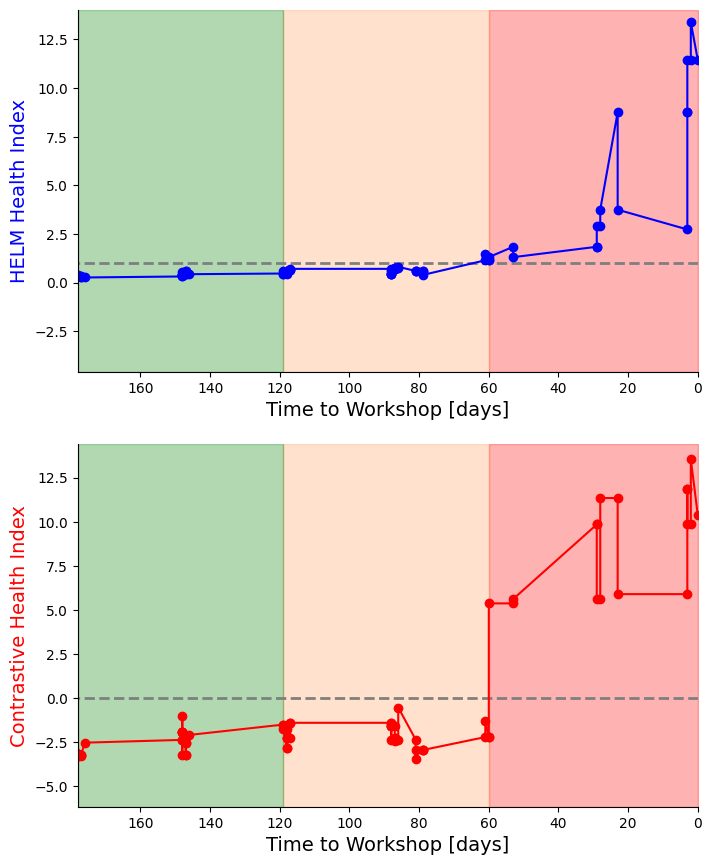}
    %\caption{Shelling Defect} 
    \label{fig:hi_shelling}}
\hfil
\subfloat[Crack defect]{
    \includegraphics[width=0.4\linewidth]{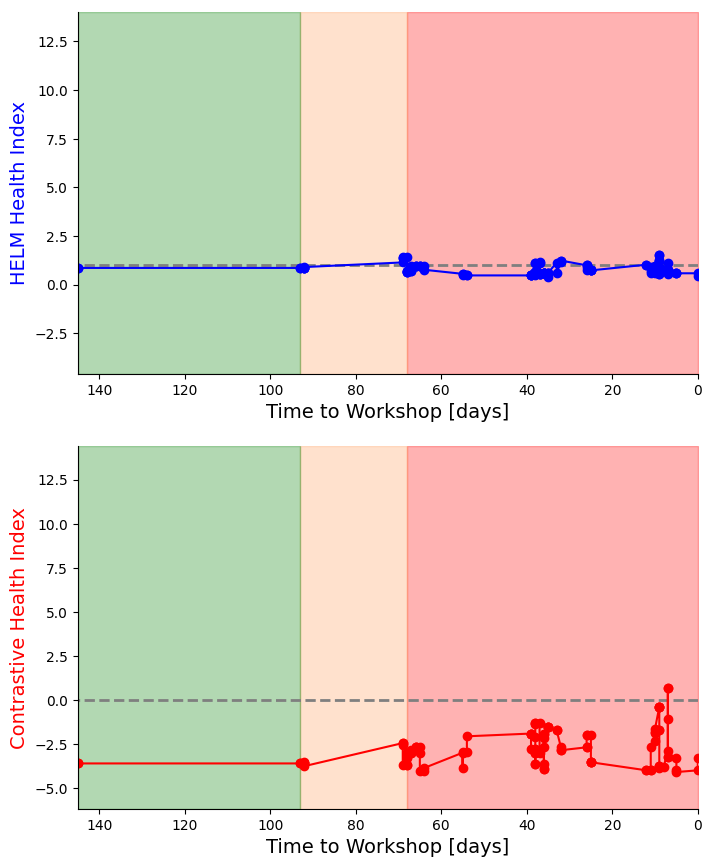}
    %\caption{Crack Defect}
    \label{fig:hi_cracks}}
\caption{Health indicator of selected wheel trajectories before the workshop visit. The \textit{dynCoeff} is plotted in green on top, HELM is shown in blue in the middle and the {\color{black}C}ontrastive model is shown in red at the bottom.}
\label{fig:hi}
\end{figure*}

{\color{black}As the ensemble model results in the highest performance, w}e present two examples of health monitoring over time {\color{black}of the two involved models of the ensemble}: one for a wheel with shelling defects (see \figref{fig:hi_shelling}) and another for a wheel  with crack defects (see \figref{fig:hi_cracks}). Due to the partial observability of this case study, the health indicator only appears to recover but in fact, some measurements might simply not capture the faulty part of the wheel. Therefore, the health indicator based on this partially observable data is neither smooth nor monotone since measurements that did not capture a local defect, correspond to healthy measurements. This reflects correctly in the health indicator that shows healthy conditions. 

The background color in \figref{fig:hi_shelling} and \figref{fig:hi_cracks} indicates the ground truth label  assigned by domain experts, as defined in \figref{fig:annotation}. The x-axis represents  the number of days remaining  until the wheel was inspected and the defect was identified. The HELM health ind{\color{black}icator} is scaled by the threshold proposed in \cite{michau2020feature}, while the contrastive health ind{\color{black}icator} is scaled by a constant value of 50. In the visualizations,  the HELM health ind{\color{black}icator} is {\color{black}plotted on the top} (blue line) and  the health ind{\color{black}icator} extracted from the contrastive feature space is shown at the bottom (red line). 
%The \textit{dynCoeff} fails to detect both the shelling and crack defects. 
Both methodologies can detect the shelling defect, even at the same point in time. However, for the {\color{black}C}ontrastive model at the bottom,  a clear jump is visible at an early stage, suggesting that the learned feature representation is sensitive to variations in the data caused by shelling defects. For the crack defect, both models show less sensitivity. However, the HELM model exhibits higher sensitivity, as it detects the defect considerably earlier (in the orange zone). 

\textbf{Shelling Defect Detection Time:}
The detection times of {\color{black}the best performing models and the model currently used in production} are evaluated in \tabref{tab:time}, where the  model's detection times are compared with the domain experts' annotations, as described in \secref{sec:ExpSetUP}. 
The model based on the \textit{dynCoeff} detected 3 out of 26 shelling defects (see TP column), and all of these detections occurred  after the fault became obvious in the data (in the red zone), shortly before the next workshop visit (with $dr$ close to 1). %{\color{black} The OC-SVM model detects 21 out of the shelling defects and the detection time is mainly in the orange zone (12 wheels) or in the red zone (7 wheels). Equally,} 
HELM detected 21 out of the 26 shelling defects, with the majority (10 wheels) being detected in the green zone, before the defect became obvious in the data. {\color{black}Nine wheels are detected within the orange zone as exemplified on one wheel trajectory in \figref{fig:hi_shelling}}. The two wheels {\color{black} that are} detected in the red zone {\color{black} by the HELM model are}  still identified  close to the expert's label ($dr<0.5$). The {\color{black}C}ontrastive model detect{\color{black}s} most of the shelling defects (23 out of 26 defective wheels), with the majority being detected in the orange zone (15 wheels).
\begin{table*}[h]

\caption{Time of railway wheel fault detection on shelling and crack defects. Column 1 shows the defects that were falsely labeled as healthy (FN) and the true positive (TP) detected defective wheels. Column 2 shows the total number of correctly labeled defective (TP) wheels that were labeled in the green ($dt<0$),  orange ($dt=0$), or red ($dt>0$) zone in \figref{fig:annotation}.  Column 3 shows the relative value of the time difference to the total time of the defect (DT) of the wheels detected in the red zone. \label{tab:time}}
\centering
\begin{tabular}{|l | cc | ccc| ccc|}
\hline
&&&\multicolumn{3}{c|}{Total Time dt [days]} &\multicolumn{3}{c|}{Relative Time $dr=\frac{dt}{DT}$}  \\ \cline{4-9}
 Method & FN & TP & $dt<0$ & $dt=0$ & $dt>0$ & $dr<0.1$ & $dr<=0.5$ & $dr>0.5$ \\ \cline{1-9}
&\multicolumn{8}{c|}{Shelling}\\\cline{1-9}
\textit{dynCoeff}. & 23 & 3  & 0 & 0 & 3 & 0 & 0& 3   \\[0.2ex]
%OC-SVM &  5&  21&  2&  12&   7&  1&  4&  2  \\[0.2ex]
HELM & 5 & 21 & 10 & 9 & 2  & 1 & 1 & 0   \\[0.2ex]
Contrastive & 3  & \textbf{23}   & 4& 15 & 4  & 1 & 1 & 2 \\ [0.2ex]\cline{1-9}
&\multicolumn{8}{c|}{Cracks}\\\cline{1-9}
\textit{dynCoeff} & 52 & 1  & 0 & 0 & 1 & 0 & 0& 1   \\[0.2ex]
%OC-SVM & 7 &  \textbf{46} & 13 & 23 & 10  & 0 & 6 &  4  \\[0.2ex]
HELM & 11 & 42 & 15 & 23 & 4  & 0 & 0 & 4  \\[0.2ex]
Contrastive & 13  & 40   & 7 & 16 & 17  & 1 & 2 & 14 \\ [0.2ex]
\hline
\end{tabular}\\[10pt]
\end{table*}

\textbf{Crack Defect Detection:}
The results for the cracks are shown in the lower half of \tabref{tab:time}. The \textit{dynCoeff} detect{\color{black}s one}  out of 53 crack defects and it {\color{black}is} detected late, shortly before the next workshop visit ($dr$ close to value 1). %{\color{black} The OC-SVM model detects most crack defects (46 out of 53) and most of them in the orange zone.} 
HELM detect{\color{black}s} 42 of the 53  crack defects, most of which were detected in the orange zone, i.e.\ exactly when the fault manifested in the data. The {\color{black}C}ontrastive model detected 40 crack defects in total, most of which were detected late (17 wheels).

\section{Discussion} \label{sec:Discussion}
In this work, we propose unsupervised contrastive feature learning to construct a robust and informative health indicator by measuring the distance to the decision boundary of a OC-SVM. % that {\color{black} is versatile in its application} and {\color{black} versatile with respect to the} downstream tasks. 
We demonstrate  how the learned health indicator captures the real degradation trend, as demonstrated on the milling machine case study where ground truth labels are available for evaluation and is suited for the downstream task of fault detection.  The key  points are discussed below.

\textbf{Contrastive Health Indicator for Condition Monitoring:} %We have demonstrated that the proposed contrastive health indicator is versatile with respect to the application and with respect to the downstream task by conducting experiments on the task of wear assessment of milling machines and on the task of fault detection for railway wheels. 
Without access to the ground truth information of the degradation state, we used the \textit{operation time} as a linear proxy for the degree of degradation {\color{black} to train the contrastive features. Thus, we} rely on the hypothesis that the monitored assets (coaches of a fleet or milling machines) are operated in a similar way and therefore, the degradation process can be assumed to be comparable in time between the different assets. 
This hypothesis only holds partially as it depends e.g.\ on the operating schedule of each individual machine or vehicle. Therefore, the \textit{operation time} is only a linear and noisy proxy for the actual degradation.
Still, the learned feature representation captures the non-linear evolution of the health condition instead of the linear evolution in time as demonstrated on the task of wear assessment on milling machines. The self-supervised selection of positive samples and the semi-hard selection seem to prevent the feature space to structure itself based on the health as reflected in the data, rather than the time. 
Furthermore, the health indicator shows not only sensitivity to degradation but can also be used to distinguish normal degradation processes from faulty conditions as demonstrated in the railway wheel case study. 
{\color{black} Compared to the OC-SVM model applied directly to the condition monitoring data (OC-SVM), the health indicator based on the {\color{black}C}ontrastive model demonstrates slightly less sensitivity to detecting defects. This indicates that the contrastive features that are learned to be invariant to certain \textit{non-informative} variations in the data within the healthy class, might be slightly less sensitive to certain fault patterns in the data. However,  it is important to emphasize that the health indicator based on the contrastive features also results in less mislabeled healthy wheels in comparison to the OC-SVM model, demonstrating that the invariance within the healthy class is useful to reduce the false alarm rate. Due to the reduced false alarm rate, the health indicator based on the contrastive features (Contrastive) outperforms the one based on the data without feature learning (OC-SVM). }

{\color{black} \textbf{Time Window Selection for Positive Sampling:} The ‘positive’ time window of the \textit{operational time} in which the health condition of an asset is not expected to change is one parameter to set for the proposed approach. In the first case study of this paper (milling machines), we have set this parameter according to the performance on a validation dataset. In the second case study, in absence of a labeled validation dataset, this value was set based on expert knowledge. While this presents some constraints (since expert knowledge is required to set this parameter in the absence of a validation dataset), we consider it realistic for practitioners with experience to determine such a timeframe during which no substantial change in the health condition is expected under a normal degradation process.}

{\color{black} \textbf{Versatility of the Proposed Approach:} Despite the differences in the two applications considered in this paper - including different modalities, tasks and characteristics with respect to observability - the contrastive features were learned in exactly the same way, and the health indicator was extracted identically from the learned features in both cases. Thus, the proposed methodology proves to be versatile. Given the significant variations in condition monitoring tasks or applications, it is necessary to anticipate that certain pre- and or post-processing steps must be customized for specific tasks. This is because the available data often comes from different modalities, is usually not standardized,  is recorded at different sampling frequencies, and might only partially observe the system. %Using signal processing techniques with high frequency data such as accelerometer data is a common pre-processing step in machine learning literature \cite{surucu2023condition, wang2020missing}. 
Similarly, HELM is versatile applicable in the same sense and it has shown higher sensitive in detecting cracks on the railway case study. On the other tasks, however, the proposed approach performed slightly better by showing more sensitivity to shelling defects and following the true health of the milling machines with a more consistent pattern (see below).} 
%Given these findings, it is crucial to evaluate which method is more suited given a specific task by validating it on more validation tasks. Since both methods are fully unsupervised, this validation process is always required in safety critical environments. We do not consider this a limitation in the versatile applicability of the approach. }

% Point 1
\textbf{Unsupervised Contrastive Feature Learning for Wear Assessment:} %Even though quantitatively, our proposed method outperforms the comparison method by a small margin based on the reported evaluation metrics (correlation, cosine similarity and trendability), 
Our method is able to follow the real evolution of the wear very closely, as becomes visible in \figref{fig:hi_milling}. 
 Particularly  during  the initial {\color{black}operation} phase of  the milling machines, our proposed health indicator closely follows the trajectory of the real wear for both machines. This suggests  that the learned feature space is very sensitive to subtle  changes in  health conditions while remaining   robust across  different monitored machines. %Further, the learned feature representation captures the non-linear evolution of the health condition.
 
Furthermore, the proposed method consistently overestimates the end-of-life wear on both milling machines by a similar margin. If this trend is confirmed by further field measurements across additional machines, identifying and correcting this systematic bias in the model could significantly improve its performance. In contrast, the health indicator learned by HELM fluctuates between overestimations and underestimations {\color{black}at the end of life and also exhibits less consistent behavior in the initial phase, notably overestimating on machine c1. This suggests} a greater sensitivity to machine-specific factors, {\color{black} as the method is based on reconstruction and thus potentially prone to small variations. Consequently,
it may be less versatile compared to the Contrastive learning approach due to its increased sensitivity to specific machines}. 
 
 One notable observation is that the initial value of the health indicator (at cycle 1) is consistently underestimated for all approaches. This is because  the training dataset is unlabeled,  making it unknown at what  wear values the represented flutes begin operating. Consequently, it is unclear whether  this low health indicator value is due to  exposure to data recorded under lower wear conditions  during training  or a challenge in generalization from the training dataset to the test dataset. Nevertheless, after the first cycle of operation, the contrastive health indicator promptly  adapts to the true wear value.  The health indicator's extrapolation ability is further discussed in the railway wheel case study (see below).

\textbf{From Degradation Sensitivity to Fault Detection:} For the task of wear assessment of milling machines, we assume that the wear values in both the training and test datasets fall  within a similar range. This implies that the model is exposed to data recorded under similar health conditions during both trai{\color{black}ni}ng and testing.  
In contrast, for the task of railway wheel fault detection, this assumption  does {\color{black} not} hold, as presumably no faults were represented in the dataset. 
In this task, the objective  of training a feature space based on degradation was to assess the extent to which an encoder model that is sensitive to degradation, also exhibits sensitivity to faults. Different fault types induce distinct  patterns in the CM data,  and the model might only be sensitive to certain patterns while displaying insensitivity to others. This  observation was also confirmed  in our experiments. Specifically, the contrastive encoder's sensitivity to degradation translates  well to sensitivity to shelling defects but not to crack defects. %The constrastive model showed the highest sensitivity to shelling defects.
%, training a feature space that is sensitive to faults without knowing is challenging given an unlabeled dataset. due to the lack of labels in the training dataset, it is unknown if faults are represented in the training dataset. Faults in the training dataset has a  By ch   the negative samples were selected as being distant in time, which potentially represents slightly degraded wheels. The features learnt in this way were well suited for detecting shelling defects. In this case, 
For shelling defects, the {\color{black}C}ontrastive model not only detected most of the defects but also accurately identified  the time of fault occurrence. However, for cracks, this  high level of  performance could not be replicated, and the proposed model detected fewer cracks compared to HELM {\color{black} and OC-SVM}. %, the contrastive encoder's sensitivity to degradation transfers well to sensitivity to shelling defects but not to crack defects. 
One possible explanation for this difference  could be that shelling defects may share  more similarities with the naturally  occurring degradation compared to cracks. Another potential  explanation is related to the absence of labels in the training dataset. It is uncertain whether faulty data was present in the training dataset. If indeed faulty data coexisted with healthy data of various degradation states in the training dataset may have created  positive pairs, making the encoder model  insensitive to the specific  fault pattern. Consequently, if the training dataset contained a significant number of crack defects, this could explain the {\color{black}C}ontrastive model's poor performance in detecting such defects. This possibility is plausible, as crack defects are generally less  visually apparent  compared to shelling defects, increasing the likelihood  that they might go unnoticed and unreported by the workshop inspectors. 
Furthermore, from a qualitative visual assessment  of the test dataset, it becomes  apparent that crack defects have a lesser impact on the strain gauge signals compared to shelling defects, making cracks more challenging  to detect with strain gauge sensors. This could  provide  another explanation for why a model trained to be invariant to certain variations in the data (caused by operating or environmental factors) exhibited  lower sensitivity to cracks. Combining both approaches {\color{black} (Contrastive + HELM)} results in the highest balanced accuracy of $88.7\%$, as the ensemble of both approaches benefits from HELM's sensitivity to cracks and the {\color{black}C}ontrastive model's  sensitivity  to shelling.

% Point 3
\textbf{Fault Detection Time Railway Wheels:} The evaluation of the detection times showed that HELM excels  in  early detection, corresponding to the green zone in \figref{fig:annotation}. Early detection is generally considered desirable as it enables early maintenance planning. However, premature detections can also lead to additional work and wasted resources if the wheel is sent to the workshop too early. Surprisingly, HELM is not the most sensitive model for all fault types, as it detected fewer shelling defects compared to the {\color{black}C}ontrastive model. Furthermore, the {\color{black}C}ontrastive model identified  most shelling defects in the same timeframe as the domain experts - detecting 15 out of 26 wheels in the orange zone. However, the contrastive learning model is less sensitive to cracks, often resulting in late detections (17 out of 53). In general, the sensitivity of the models can be adapted to the users' requirements. For instance, if the user immediately stops train operation upon detecting a defect, early detection may result in machine downtime. On the contrary, if the model is used for long-term maintenance planning, early detection is desirable. Additionally, early detection of faults and the corresponding health ind{\color{black}icator} as a severity measure can provide valuable  information, especially when fault severity has not been defined or tracked before. Early detection could offer insights  into how faults evolve, which can be verified in the future when trains  enter the depot or workshop. Consequently, the model enables monitoring of fault severity evolution over time. 

% Point 4
%\textbf{Fault Occurrence Railway Wheels:} A surprising finding in the labeling process depicted in \figref{fig:process} of the wheel defect dataset is that many of the healthy wheels in the preliminary test dataset were identified as anomalous by the domain experts, which resulted in a small number of healthy wheels in the test dataset. It should be noted that the domain experts who evaluated the data did not have access to the real condition of the wheel but only to the data. In contrast, the maintenance technicians in the workshops primarily use visual inspection information to label the health condition of the wheel. In the future, it would be interesting to investigate whether the wheels that have been detected as defective by the domain experts and the contrastive model, show a different type of defect that the maintenance technicians in the workshop might not be familiar with and may not be used to detecting through visual inspection.  

\section{Conclusion} \label{sec:Conclusion}
% Main take away
In this work, we propose to use unsupervised contrastive learning to develop  a health indicator for both fault detection and wear assessment. We demonstrate its applicability two very  different applications: wear assessment of milling machines {\color{black}where an average correlation of 0.97 is achieved with respect to the true health indicator} and health monitoring of railway wheels, where no labeled fault data was available for training the model but the training dataset was presumed to be mainly healthy. {\color{black}In this case study, a balanced accuracy of $88.7\%$ is achieved at the task of fault detection.}  %The main novelty of the proposed framework is in the unsupervised setup without any observed faults. %, where we evaluate how unsupervised contrastive learning can be implemented  with no observed faults for selecting negative samples in real applications.
%We evaluate the performance of the proposed method on two CM datasets of railway applications - one image dataset from infrastructure (sleeper) and one time-series dataset from rolling stock (wheel).
%In this research, contrastive feature learning was successfully applied to different fault detection and diagnostics tasks in the railway system. 
While  the tasks differ in many aspects, the conducted experiments demonstrated that contrastive learning improves  performance on both datasets and tasks (fault detection and wear assessment) compared to state-of-the-art methods. This supports our initial assumption that contrastive learning provides a good way to represent healthy samples and the distance from the compact representation lends itself as a good representation of the asset's health. 
In future work, we plan integrate a monoticity constraint into the health ind{\color{black}icator} and explore the suitability of the feature space for prognostics tasks.
One potentially promising direction could be to incorporate observed faults in  feature learning and investigate semi-supervised setups rather than relying  solely on unsupervised approaches. %Generalization to other fleets will also be investigated in the future. 

\section*{Acknowledgments}
This research resulted from the "Integrated intelligent railway wheel condition prediction" (INTERACT) project, supported by the ETH Mobility Initiative. Special thanks go to the domain experts from SBB, supporting our efforts with data acquisition and valuable expertise.

\ifdefined\ARXIV
\printbibliography
\fi

\ifdefined\IEEE
 \bibliographystyle{IEEEtranN}
\bibliography{References.bib}

@article{krummenacher2017wheel,
  title={Wheel defect detection with machine learning},
  author={Krummenacher, Gabriel and Ong, Cheng Soon and Koller, Stefan and Kobayashi, Seijin and Buhmann, Joachim M},
  journal={IEEE Transactions on Intelligent Transportation Systems},
  volume={19},
  number={4},
  pages={1176--1187},
  year={2017},
  publisher={IEEE}
}

@article{michau2020feature,
  title={Feature learning for fault detection in high-dimensional condition monitoring signals},
  author={Michau, Gabriel and Hu, Yang and Palm{\'e}, Thomas and Fink, Olga},
  journal={Proceedings of the Institution of Mechanical Engineers, Part O: Journal of Risk and Reliability},
  volume={234},
  number={1},
  pages={104--115},
  year={2020},
  publisher={SAGE Publications Sage UK: London, England}
}

@inproceedings{chopra2005learning,
  title={Learning a similarity metric discriminatively, with application to face verification},
  author={Chopra, Sumit and Hadsell, Raia and LeCun, Yann},
  booktitle={2005 IEEE Computer Society Conference on Computer Vision and Pattern Recognition (CVPR'05)},
  volume={1},
  pages={539--546},
  year={2005},
  organization={IEEE}
}

@inproceedings{schroff2015facenet,
  title={Facenet: A unified embedding for face recognition and clustering},
  author={Schroff, Florian and Kalenichenko, Dmitry and Philbin, James},
  booktitle={Proceedings of the IEEE conference on computer vision and pattern recognition},
  pages={815--823},
  year={2015}
}

@inproceedings{chen2020simple,
  title={A simple framework for contrastive learning of visual representations},
  author={Chen, Ting and Kornblith, Simon and Norouzi, Mohammad and Hinton, Geoffrey},
  booktitle={International conference on machine learning},
  pages={1597--1607},
  year={2020},
  organization={PMLR}
}

@article{franceschi2019unsupervised,
  title={Unsupervised scalable representation learning for multivariate time series},
  author={Franceschi, Jean-Yves and Dieuleveut, Aymeric and Jaggi, Martin},
  journal={Advances in neural information processing systems},
  volume={32},
  year={2019}
}

@article{rombach2021contrastive,
  title={Contrastive Learning for Fault Detection and Diagnostics in the Context of Changing Operating Conditions and Novel Fault Types},
  author={Rombach, Katharina and Michau, Gabriel and Fink, Olga},
  journal={Sensors},
  volume={21},
  number={10},
  pages={3550},
  year={2021},
  publisher={Multidisciplinary Digital Publishing Institute}
}

@article{zhou2022construction,
  title={Construction of health indicators for condition monitoring of rotating machinery: A review of the research},
  author={Zhou, Haoxuan and Huang, Xin and Wen, Guangrui and Lei, Zihao and Dong, Shuzhi and Zhang, Ping and Chen, Xuefeng},
  journal={Expert Systems with Applications},
  volume={203},
  pages={117297},
  year={2022},
  publisher={Elsevier}
}

@inproceedings{li2009fuzzy,
  title={Fuzzy neural network modelling for tool wear estimation in dry milling operation},
  author={Li, Xiang and Lim, BS and Zhou, JH and Huang, S and Phua, SJ and Shaw, KC and Er, MJ},
  booktitle={Annual Conference of the PHM Society},
  volume={1},
  number={1},
  year={2009}
}

@article{li2022intelligent,
  title={Intelligent tool wear prediction based on Informer encoder and stacked bidirectional gated recurrent unit},
  author={Li, Wangyang and Fu, Hongya and Han, Zhenyu and Zhang, Xing and Jin, Hongyu},
  journal={Robotics and Computer-Integrated Manufacturing},
  volume={77},
  pages={102368},
  year={2022},
  publisher={Elsevier}
}

@article{liu2022three,
  title={Three-stage wiener-process-based model for remaining useful life prediction of a cutting tool in high-speed milling},
  author={Liu, Weichao and Yang, Wen-An and You, Youpeng},
  journal={Sensors},
  volume={22},
  number={13},
  pages={4763},
  year={2022},
  publisher={MDPI}
}

@article{he2022milling,
  title={Milling tool wear prediction using multi-sensor feature fusion based on stacked sparse autoencoders},
  author={He, Zhaopeng and Shi, Tielin and Xuan, Jianping},
  journal={Measurement},
  volume={190},
  pages={110719},
  year={2022},
  publisher={Elsevier}
}

@article{de2023developing,
  title={Developing health indicators and RUL prognostics for systems with few failure instances and varying operating conditions using a LSTM autoencoder},
  author={de Pater, Ingeborg and Mitici, Mihaela},
  journal={Engineering Applications of Artificial Intelligence},
  volume={117},
  pages={105582},
  year={2023},
  publisher={Elsevier}
}

@article{fink2020potential,
  title={Potential, challenges and future directions for deep learning in prognostics and health management applications},
  author={Fink, Olga and Wang, Qin and Svensen, Markus and Dersin, Pierre and Lee, Wan-Jui and Ducoffe, Melanie},
  journal={Engineering Applications of Artificial Intelligence},
  volume={92},
  pages={103678},
  year={2020},
  publisher={Elsevier}
}

@article{garmaev2023deep,
  title={Deep Koopman Operator-based degradation modelling},
  author={Garmaev, Sergei and Fink, Olga},
  journal={arXiv preprint arXiv:2308.01690},
  year={2023}
}

@article{atamuradov2020machine,
  title={Machine health indicator construction framework for failure diagnostics and prognostics},
  author={Atamuradov, Vepa and Medjaher, Kamal and Camci, Fatih and Zerhouni, Noureddine and Dersin, Pierre and Lamoureux, Benjamin},
  journal={Journal of signal processing systems},
  volume={92},
  pages={591--609},
  year={2020},
  publisher={Springer}
}

@article{lei2018machinery,
  title={Machinery health prognostics: A systematic review from data acquisition to RUL prediction},
  author={Lei, Yaguo and Li, Naipeng and Guo, Liang and Li, Ningbo and Yan, Tao and Lin, Jing},
  journal={Mechanical systems and signal processing},
  volume={104},
  pages={799--834},
  year={2018},
  publisher={Elsevier}
}

@article{ye2021health,
  title={Health condition monitoring of machines based on long short-term memory convolutional autoencoder},
  author={Ye, Zhuang and Yu, Jianbo},
  journal={Applied Soft Computing},
  volume={107},
  pages={107379},
  year={2021},
  publisher={Elsevier}
}

@article{ni2022data,
  title={Data-driven prognostic scheme for bearings based on a novel health indicator and gated recurrent unit network},
  author={Ni, Qing and Ji, JC and Feng, Ke},
  journal={IEEE Transactions on Industrial Informatics},
  volume={19},
  number={2},
  pages={1301--1311},
  year={2022},
  publisher={IEEE}
}

@article{liu2020complex,
  title={Complex engineered system health indexes extraction using low frequency raw time-series data based on deep learning methods},
  author={Liu, Cui and Sun, Jianzhong and Liu, He and Lei, Shiying and Hu, Xinhua},
  journal={Measurement},
  volume={161},
  pages={107890},
  year={2020},
  publisher={Elsevier}
}

@article{malhotra2016multi,
  title={Multi-sensor prognostics using an unsupervised health index based on LSTM encoder-decoder},
  author={Malhotra, Pankaj and Tv, Vishnu and Ramakrishnan, Anusha and Anand, Gaurangi and Vig, Lovekesh and Agarwal, Puneet and Shroff, Gautam},
  journal={arXiv preprint arXiv:1608.06154},
  year={2016}
}

@article{fu2021novel,
  title={A novel time-series memory auto-encoder with sequentially updated reconstructions for remaining useful life prediction},
  author={Fu, Song and Zhong, Shisheng and Lin, Lin and Zhao, Minghang},
  journal={IEEE Transactions on Neural Networks and Learning Systems},
  volume={33},
  number={12},
  pages={7114--7125},
  year={2021},
  publisher={IEEE}
}

@article{zhai2021enabling,
  title={Enabling predictive maintenance integrated production scheduling by operation-specific health prognostics with generative deep learning},
  author={Zhai, Simon and Gehring, Benedikt and Reinhart, Gunther},
  journal={Journal of Manufacturing Systems},
  volume={61},
  pages={830--855},
  year={2021},
  publisher={Elsevier}
}

@article{michau2022fully,
  title={Fully learnable deep wavelet transform for unsupervised monitoring of high-frequency time series},
  author={Michau, Gabriel and Frusque, Gaetan and Fink, Olga},
  journal={Proceedings of the National Academy of Sciences},
  volume={119},
  number={8},
  pages={e2106598119},
  year={2022},
  publisher={National Acad Sciences}
}

@article{stetco2019machine,
  title={Machine learning methods for wind turbine condition monitoring: A review},
  author={Stetco, Adrian and Dinmohammadi, Fateme and Zhao, Xingyu and Robu, Valentin and Flynn, David and Barnes, Mike and Keane, John and Nenadic, Goran},
  journal={Renewable energy},
  volume={133},
  pages={620--635},
  year={2019},
  publisher={Elsevier}
}

@article{zhang2023health,
  title={Health indicator based on signal probability distribution measures for machinery condition monitoring},
  author={Zhang, Guangyao and Wang, Yi and Li, Xiaomeng and Qin, Yi and Tang, Baoping},
  journal={Mechanical Systems and Signal Processing},
  volume={198},
  pages={110460},
  year={2023},
  publisher={Elsevier}
}

@article{kong2023contrastive,
  title={A contrastive learning framework enhanced by unlabeled samples for remaining useful life prediction},
  author={Kong, Ziqian and Jin, Xiaohang and Xu, Zhengguo and Chen, Zian},
  journal={Reliability Engineering \& System Safety},
  volume={234},
  pages={109163},
  year={2023},
  publisher={Elsevier}
}

@article{michau2021unsupervised,
  title={Unsupervised transfer learning for anomaly detection: Application to complementary operating condition transfer},
  author={Michau, Gabriel and Fink, Olga},
  journal={Knowledge-Based Systems},
  volume={216},
  pages={106816},
  year={2021},
  publisher={Elsevier}
}

@article{guo2018machinery,
  title={Machinery health indicator construction based on convolutional neural networks considering trend burr},
  author={Guo, Liang and Lei, Yaguo and Li, Naipeng and Yan, Tao and Li, Ningbo},
  journal={Neurocomputing},
  volume={292},
  pages={142--150},
  year={2018},
  publisher={Elsevier}
}

@article{lu2014intelligent,
  title={An intelligent approach to machine component health prognostics by utilizing only truncated histories},
  author={Lu, Chen and Tao, Laifa and Fan, Huanzhen},
  journal={Mechanical Systems and Signal Processing},
  volume={42},
  number={1-2},
  pages={300--313},
  year={2014},
  publisher={Elsevier}
}

@article{firla2016automatic,
  title={Automatic characteristic frequency association and all-sideband demodulation for the detection of a bearing fault},
  author={Firla, Marcin and Li, Zhong-Yang and Martin, Nadine and Pachaud, Christian and Barszcz, Tomasz},
  journal={Mechanical Systems and Signal Processing},
  volume={80},
  pages={335--348},
  year={2016},
  publisher={Elsevier}
}

@article{saidi2017wind,
  title={Wind turbine high-speed shaft bearings health prognosis through a spectral Kurtosis-derived indices and SVR},
  author={Saidi, Lotfi and Ali, Jaouher Ben and Bechhoefer, Eric and Benbouzid, Mohamed},
  journal={Applied Acoustics},
  volume={120},
  pages={1--8},
  year={2017},
  publisher={Elsevier}
}

@article{wang2017prognostics,
  title={Prognostics and health management: A review of vibration based bearing and gear health indicators},
  author={Wang, Dong and Tsui, Kwok-Leung and Miao, Qiang},
  journal={Ieee Access},
  volume={6},
  pages={665--676},
  year={2017},
  publisher={IEEE}
}

@article{chen2020anomaly,
  title={Anomaly detection for wind turbines based on the reconstruction of condition parameters using stacked denoising autoencoders},
  author={Chen, Junsheng and Li, Jian and Chen, Weigen and Wang, Youyuan and Jiang, Tianyan},
  journal={Renewable Energy},
  volume={147},
  pages={1469--1480},
  year={2020},
  publisher={Elsevier}
}

@article{gonzalez2022health,
  title={Health indicator for machine condition monitoring built in the latent space of a deep autoencoder},
  author={Gonz{\'a}lez-Mu{\~n}iz, Ana and Diaz, Ignacio and Cuadrado, Abel A and Garc{\'\i}a-P{\'e}rez, Diego},
  journal={Reliability Engineering \& System Safety},
  volume={224},
  pages={108482},
  year={2022},
  publisher={Elsevier}
}

@article{sadooghi2018improving,
  title={Improving one class support vector machine novelty detection scheme using nonlinear features},
  author={Sadooghi, Mohammad Saleh and Khadem, Siamak Esmaeilzadeh},
  journal={Pattern Recognition},
  volume={83},
  pages={14--33},
  year={2018},
  publisher={Elsevier}
}

@article{wang2018theoretical,
  title={Theoretical investigation of the upper and lower bounds of a generalized dimensionless bearing health indicator},
  author={Wang, Dong and Tsui, Kwok-Leung},
  journal={Mechanical systems and signal processing},
  volume={98},
  pages={890--901},
  year={2018},
  publisher={Elsevier}
}

@misc{hsu2023comparison,
      title={A Comparison of Residual-based Methods on Fault Detection},
      author={Chi-Ching Hsu and Gaetan Frusque and Olga Fink},
      year={2023},
      eprint={2309.02274},
      archivePrefix={arXiv},
      primaryClass={eess.SY}
}

@article{song2017integration,
  title={Integration of data-level fusion model and kernel methods for degradation modeling and prognostic analysis},
  author={Song, Changyue and Liu, Kaibo and Zhang, Xi},
  journal={IEEE Transactions on Reliability},
  volume={67},
  number={2},
  pages={640--650},
  year={2017},
  publisher={IEEE}
}

@article{chen2022deep,
  title={A deep learning feature fusion based health index construction method for prognostics using multiobjective optimization},
  author={Chen, Zhen and Zhou, Di and Zio, Enrico and Xia, Tangbin and Pan, Ershun},
  journal={IEEE Transactions on Reliability},
  year={2022},
  publisher={IEEE}
}

@article{rail,
  title={Increasing the wheelset life by understanding the wheel conditions},
  author={Vincent, David and Lager, Rolf},
  journal={ZEVrail},
  year={2021},
  publisher={Siemens}
}
\fi

\vfill

\end{document}